\documentclass[lettersize,journal]{IEEEtran}
\usepackage{amsmath,amsfonts}
\usepackage{algorithmic}
\usepackage{algorithm}
\usepackage{array}
\usepackage[caption=false,font=normalsize,labelfont=sf,textfont=sf]{subfig}
\usepackage{textcomp}
\usepackage{stfloats}
\usepackage{url}
\usepackage{verbatim}
\usepackage{graphicx}
\usepackage{cite}
\usepackage{bbding}
\usepackage{pifont}
\usepackage{wasysym}
\usepackage{multicol}
\usepackage{multirow}
\usepackage{makecell}
\usepackage{bm}
\usepackage{balance}
\usepackage{booktabs}
\usepackage{tablefootnote}
\usepackage{color, soul}

\usepackage[table,xcdraw]{xcolor}  

\arrayrulecolor{black}  

\newcommand{\blue}{\textcolor[rgb]{0.00,0.00,0.00}}

\begin{document}

\title{UP-Person: Unified Parameter-Efficient Transfer Learning for Text-based Person Retrieval}

\author{
       Yating Liu, Yaowei Li,
        Xiangyuan Lan,
        
        Wenming Yang, \IEEEmembership{Senior Member, IEEE},
        Zimo Liu, 
        and Qingmin Liao, \IEEEmembership{Senior Member, IEEE}
\thanks{
Manuscript initially received on May 2, 2024. Revised version received on October 16, 2024, January 31, 2025. 
This work was supported by the National Natural Science Foundation of China NSFC under U23B2030, and the Major Key Project of Peng Cheng Laboratory under PCL2023A08.
(Corresponding author: Zimo Liu and Qingmin Liao.)

Yating Liu is with Shenzhen International Graduate School, Tsinghua University, Shenzhen 518071, China and Peng Cheng Laboratory, Shenzhen 518071, China (e-mail: liuyatin21@mails.tsinghua.edu.cn).

Yaowei Li is with School of ECE, Peking University, Shenzhen 518071, China and Peng Cheng Laboratory, Shenzhen 518071, China (e-mail: ywl@stu.pku.edu.cn).

Xiangyuan Lan and Zimo Liu are with Peng Cheng Laboratory, Shenzhen 518071, China (e-mail: lanxy@pcl.ac.cn; liuzm@pcl.ac.cn).

Wenming Yang and Qingmin Liao are with Shenzhen International Graduate School, Tsinghua University, Shenzhen 518071, China (e-mail: yang.wenming@sz.tsinghua.edu.cn; liaoqm@tsinghua.edu.cn).
}

}


\markboth{IEEE TRANSACTIONS ON CIRCUITS AND SYSTEMS FOR VIDEO TECHNOLOGY}%
{Shell \MakeLowercase{\textit{et al.}}: A Sample Article Using IEEEtran.cls for IEEE Journals}


\maketitle

\begin{abstract}
Text-based Person Retrieval (TPR) as a multi-modal task,
which aims to retrieve the target person from a pool of candidate images given a text description,
has recently garnered considerable attention due to the progress of contrastive visual-language pre-trained model. 
Prior works leverage pre-trained CLIP to extract person visual and textual features and fully fine-tune the entire network, which have shown notable performance improvements compared to uni-modal pre-training models.
However, full-tuning a large model is prone to overfitting and hinders the generalization ability. 
In this paper, we propose a novel \emph{U}nified \emph{P}arameter-Efficient Transfer Learning (PETL) method for Text-based \emph{Person} Retrieval (UP-Person) to thoroughly transfer the multi-modal knowledge from CLIP.
Specifically, UP-Person simultaneously integrates three lightweight PETL components including Prefix, LoRA and Adapter, where Prefix and LoRA are devised together to mine local information with task-specific information prompts, and Adapter is designed to adjust global feature representations.
Additionally, two vanilla submodules are optimized to adapt to the unified architecture of TPR.
For one thing, S-Prefix is proposed to boost attention of prefix and enhance the gradient propagation of prefix tokens, which improves the flexibility and performance of the vanilla prefix.
For another thing, L-Adapter is designed in parallel with layer normalization to adjust the overall distribution, which can resolve conflicts caused by overlap and interaction among multiple submodules.
Extensive experimental results demonstrate that our UP-Person achieves state-of-the-art results across various person retrieval datasets, including CUHK-PEDES, ICFG-PEDES and RSTPReid while merely fine-tuning 4.7\% parameters.
\blue{Code is available at https://github.com/Liu-Yating/UP-Person}.

\end{abstract}


\section{Introduction}

\IEEEPARstart{T}{ext}-based Person Retrieval (TPR) \cite{Li_2017_CVPR}  aims to locate the person of interest from a large pool of candidates given a pedestrian description, which is a cross-task that integrates person re-identification (Re-ID) \cite{chen2024multi} with cross-modal retrieval \cite{li2024fast}.
The core of TPR is to establish the matching relationship between person images and texts.
Compared to conventional image-based person retrieval (Re-ID) \cite{chen2024multi,nikhal2024hashreid,xu2022rank,yuan2023searching} and attribute-based person retrieval \cite{cormier2024upar}, text-based person retrieval \cite{niu2024overview} provides a more intuitive and convenient way by forming queries with natural language descriptions, thus attracts increasing attention from both academia and industry, benefiting a variety of applications, such as security surveillance and intelligent transportation.

\begin{figure}[!t]
  \centerline{\includegraphics[scale=0.28]{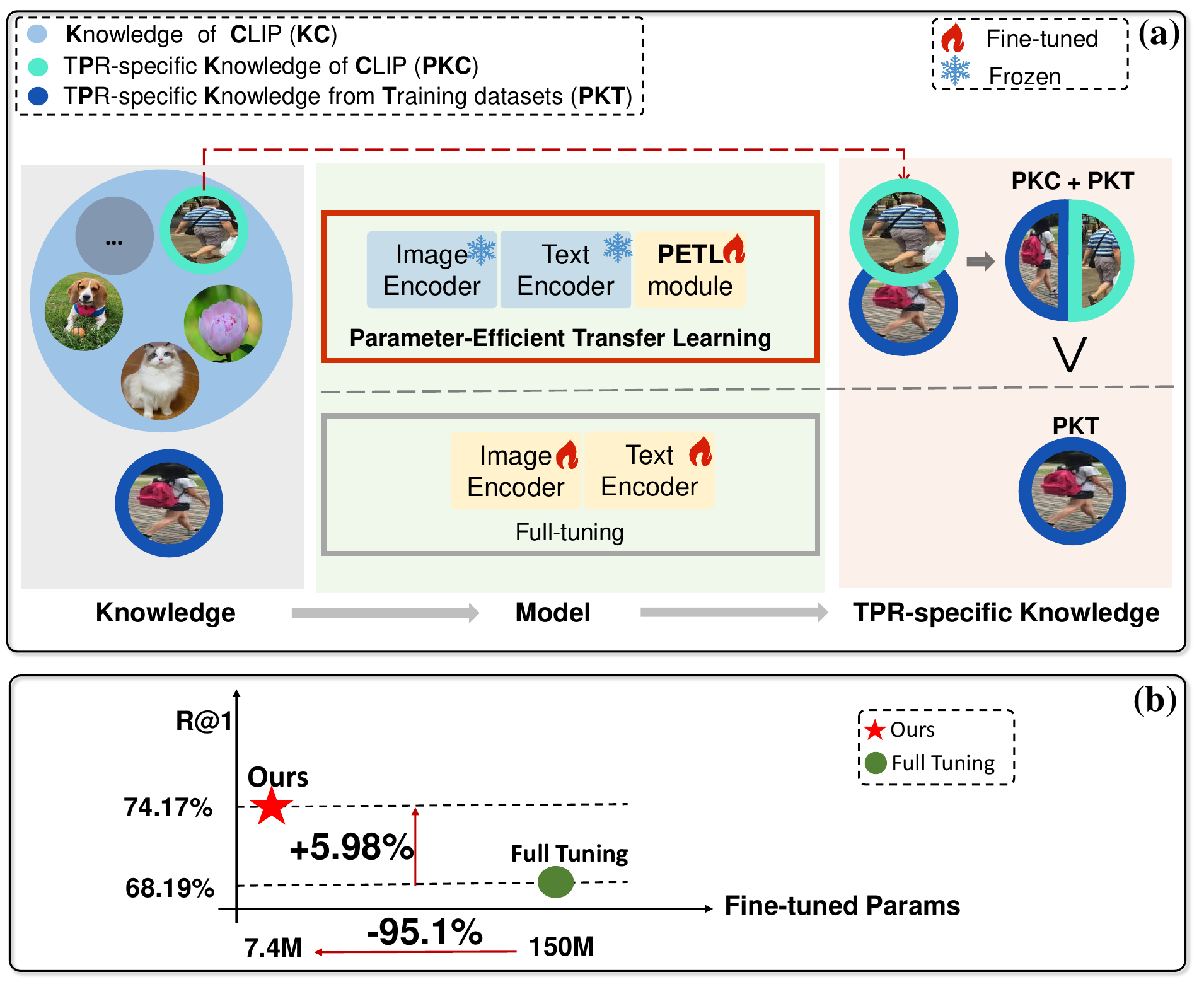}}
  \vspace{-2mm}
  \caption{\textbf{The motivation for our proposed method.}
(a) shows how PETL-based methods can transfer TPR-specific knowledge from both CLIP and training data, whereas full-tuning relies solely on the training data as its knowledge source.
Full-tuning (\emph{lower}) only utilizes the knowledge of the pre-trained CLIP at initialization and almost loses the original knowledge, which thus only retains the knowledge of TPR from training datasets (PKT).
PETL (\emph{upper}) fine-tunes a small parameters and keeps CLIP backbone frozen, which means that it can integrate both TPR-specific within CLIP (PKC) due to the retained parameters of original CLIP and TPR-specific knowledge from training data (PKT). 
Therefore, PETL methods can incorporate more knowledge compared to full-tuning if rationally designed. 
(b) On CUHK-PEDES, our approach reduces 95.1\% training parameters and gains an improvement by 5.98\% on R@1 compared to the full-tuning CLIP.}

\label{fig:motivation}
\vspace{-6mm}
\end{figure}

As large foundation models expand, the embedded knowledge becomes increasingly abundant.
Consequently, it is crucial to explore how to effectively transfer these pre-trained models to downstream tasks in order to maximize knowledge utilization.
The great success in Vision-Language Pre-training (VLP) has recently shown a strong cross-modal transfer capability in various vision-language understanding tasks \cite{zhang2024vision} , where the most representative work is Contrastive Language-Image Pre-training (CLIP) \cite{clip}.
With 400M web-crawled trainable image-text pairs, CLIP contains abundant generic knowledge learned from the large-scale dataset.
Besides, the pre-trained encoders of CLIP have greater cross-modal matching potential due to its two-branch contrastive architecture than uni-modal encoders, \emph{i.e.}, ViT \cite{dosovitskiy2020image} for vision, BERT \cite{devlin2018bert} for language.
Therefore, several works \cite{jiang2023cross,yan2023clip} adopt CLIP as the backbone, and propose multi-level matching modules to achieve CLIP-based cross-modal interactions from coarse to fine with fully fine-tuning, which leads to a significant improvement compared to many previous uni-modal retrieval frameworks \cite{shu2022see,chen2022tipcb,farooq2022axm,wu2021lapscore}.
However, this full-tuning paradigm faces two fatal issues: (1) it has a high risk of overfitting on limited task-specific training datasets as the scale of large pre-trained models continues to increase, and (2) training and storing a completely new large model for each dataset independently is expensive in practice.

An elegant solution to the above-mentioned problems is Parameter-Efficient Transfer Learning (PETL) \cite{han2024parameter}, \emph{i.e.}, Adapter \cite{gao2023clip}, LoRA \cite{hu2021LoRA   }, Prefix \cite{li2021prefix}, Prompt \cite{liu2021p} and other related variants \cite{zhang2021tip,chavan2023one,zhou2022learning}, which can achieve comparable or superior performance only with fine-tuning a few parameters of large models.
As shown in Figure \ref{fig:motivation}(a), full-tuning (\emph{lower}) only utilizes the knowledge of the pre-trained CLIP at initialization stage and almost loses the original knowledge embedded in network.
On the contrary, the CLIP backbone is frozen and preserved in PETL paradigm (\emph{upper}). 
Guided by PETL-related components, TPR-specific knowledge within CLIP is effectively transferred as the parameters of the original CLIP remain unchanged. 
The rich knowledge of CLIP about visual objects and textual descriptions can serve as knowledge complementarity for fine-grained TPR task.
Consequently, PETL facilitates learning both task-specific knowledge from general CLIP and knowledge from the training dataset, making it superior to full-tuning in terms of knowledge retention,  particularly in our scenarios with limited training data.

However, PETL paradigm is not well explored in TPR.
CSKT \cite{liu2024clip} makes the first attempt to explore CLIP with PETL-related methods on TPR and proposes a novel bidirectional multi-modal prompt-tuning, which attains superior performance only with fine-tuning 7.4\% parameters of CLIP. 
This existing PETL-based method only focuses on the global feature representation, and has not sufficiently transferred both global and local information from multiple views of CLIP.
The depth of exploration by PETL on TPR still remains limited.
This prompts us to consider whether it is feasible to design a unified PETL framework for TPR from a more comprehensive perspective.

In this paper, our target is to investigate how to design a \textbf{simple, effective and parameter-efficient \emph{unified}} transfer learning architecture based on multiple lightweight PETL methods. 
A serious concern on unifying various PETL methods is component conflict, where the performance drops significantly in practice when a single PETL component is incorporated into a unified framework. 
This occurs as the original structures of multiple PETL components overlap and interact, potentially disrupting the intended optimization direction. 
Consequently, it is crucial to design a unified framework where different components do not interfere with each other and can work cooperatively.
Another issue we observe is that when the vanilla prefix component is adapted to TPR, it shows a poor performance compared to the approximate PETL submodule such as prompt-tuning in CSKT \cite{liu2024clip}.
Thus, we consider whether we could optimize the vanilla PETL components for better synergy in the unified framework.

To address the aforementioned issues, we propose a novel \textbf{U}nified \textbf{P}arameter-Efficient Transfer Learning (PETL) method for Text-based \textbf{Person} Retrieval \textbf{(UP-Person)}. 
UP-Person implements a comprehensive PETL-based method to enhance knowledge transferring without requiring any additional complex cross-modal interaction modules. 
Specifically, as shown in Figure \ref{fig:framework}, in a transformer block of CLIP, we design and unify three submodules based on Prefix, LoRA  and Adapter to adapt to TPR task.
LoRA is incorporated  to modify the weights of Multi-Head Attention (MHA), allowing the model to capture more nuanced and local features and relationships in TPR that the original CLIP cannot fully uncover.
Prefix is prepended to the keys and values of MHA to enrich  task-specific information for TPR.
Adapter is designed in layer normalization (layernorm) to adjust the overall distribution from a global perspective.
Meanwhile, we propose two improved submodules: Salable Prefix (S-Prefix) and Layernorm Adapter (L-Adapter).
S-Prefix is introduced to enhance gradient backward propagation of prefix embeddings, which improves both the flexibility and performance of the vanilla prefix. 
L-Adapter is designed in parallel with layernorm, alongside the residual connection, to get rid of component conflicts.
Our method demonstrates greater advantages in data-scarce scenarios especially in RSTPReid dataset.
It achieves better performance and parameter-efficiency with negligible sacrifice in inference efficiency, and reduces computation and storage costs.
Our contributions can be summarized as follows:
    \begin{itemize}
        \item[$\bullet$]
        We propose a unified parameter-efficient transfer learning method for text-based person retrieval. 
        To the best of our knowledge, our study is the first attempt to investigate the unified PETL framework for TPR, which effectively transfers both  \textbf{global and local knowledge}, along with \textbf{task-specific knowledge}, to TPR task with very \textbf{fewer computation and storage costs}.
        \item[$\bullet$]
        To optimize the PETL components for better synergy, we further develop two improved PETL components, respectively: (1) a scalable prefix in attention named \textbf{S-Prefix}, and (2) a novel adapter \textbf{L-Adapter} designed in parallel with layernorm of blocks. 
        \item[$\bullet$] Extensive experiments show that UP-Person achieves superior performance compared with the prior state-of-the art on three public datasets while merely fine-tuning 4.7\% parameters.
    \end{itemize}
    
\section{Related Work}

In this section, we will briefly review the most relevant study including vision-language pre-training, parameter-efficient transfer learning, and text-based person retrieval.
\subsection{Vision-Language Pre-training.}
Vision-Language Pre-training (VLP), incorporating both image-text and video-text pre-training, focuses on learning semantic correspondence between heterogeneous modalities through pre-training on large datasets.
Generally, VLP models include five modules: vision encoder, text encoder, multimodal fusion, decoder (optional) and pre-training objective.

For vision encoder, VLP models in recent works \cite{clip,li2021align,li2023blip} adopt a pre-trained vision transformer to encode patch embeddings, such as ViT \cite{dosovitskiy2020image} or swin transformer \cite{liu2021swin}.
For language, textual pre-trained transformer like BERT \cite{devlin2018bert} can be utilized to encode word embeddings.
Furthermore, models can be categorized into two types from the multi-modal fusion perspective: dual-stream and single-stream.
For former, text and visual features are independently encoded by two transformer branches \cite{clip,li2021align,li2023blip}.
The latter single-stream models \cite{khan2022single,zhou2020unified} concatenate the text and visual features and then input them into a single shared transformer structure, which have less parameters compared with dual-stream models.
In addition to VLP models that solely employ an encoder, the encoder-decoder architecture \cite{zhou2020unified} incorporates a decoder structure, which feeds the representations into a encoder first and then sends outputs to an decoder.
Decoder is extremely helpful when combining other generative tasks such as image captioning and Visual Question Answering (VQA).
The VLP models are trained in an end-to-end manner under the guidance of the pre-training objectives \cite{zhang2024vision}, such as Masked Language Modeling (MLM), Next Sequence Prediction (NSP), Masked Vision Modeling (MVM), Image-Text Matching (ITM), Image-Text Contrastive (ITC), Word-Region Alignment (WGA) and other task-specific objectives.

In addition to model architecture, the huge training datasets are crucial to performance improvement of VLP models.
The common pre-training datasets like LAION-400M \cite{schuhmann2021laion} and LAION-5B \cite{schuhmann2022laion} are mainly composed of numerous public image-caption pairs, visual questions and answers, or larger crawled data from internet.
The abundant knowledge embedded in VLP can be applied to a broad range of downstream tasks including  cross-modal classification, regression, generation and retrieval.
In this work, we focus on text-based person retrieval based on vision-language pre-training model CLIP (400M trainable image-text pairs) due to its general knowledge, and aim to bridge the gap between pre-training large models and downstream tasks by transfer learning.

\subsection{Parameter-Efficient Transfer Learning.}
With the rapid advancement of large models,  parameter-efficient transfer learning (PETL) \cite{han2024parameter} has gained substantial attention from researchers . 
In this work, we focus on leveraging vision-language models to effectively transfer and enhance knowledge for the downstream TPR task, with a particular emphasis on PETL techniques.

As the size of pre-trained models continues to increase, from language models like BERT \cite{devlin2018bert} with 110 million parameters to  Yi-VL-34B \cite{young2024yi} with 34 billion parameters and llama 3 \cite{dubey2024llama} with 70 billion parameters, full fine-tuning will be more and more time-consuming, computationally expensive and storage-inefficient, especially for multiple domains.
PETL has emerged as a viable strategy to compensate for the above disadvantages of full-tuning, which can be broadly categorized into three types \cite{han2024parameter}: additive fine-tuning, reparameterized fine-tuning and  selective fine-tuning.
\textbf{Additive} fine-tuning methods such as Adapter \cite{houlsby2019parameter}, Prompt \cite{liu2021p} and Prefix \cite{li2021prefix} were initially introduced to facilitate the transfer of large language models to specific downstream tasks by inserting additional parameters to models.
LoRA   \cite{hu2021LoRA} as a representative method for \textbf{reparameterized} fine-tuning methods, utilizes low-rank decomposition to reconstruct the weight matrices.
\textbf{Selective} fine-tuning aims to reduce the number of fine-tuned parameters by selecting a subset of pre-trained parameters, such as Bitfit \cite{zaken2021bitfit}.

Subsequently, inspired by PETL in NLP, solutions like VPT \cite{jia2022visual} and AdapterFormer \cite{chen2022adaptformer} have emerged to address challenges in vision transfer learning.
With the success of VLP \cite{zhang2024vision}, PETL on VLP becomes a novel trend.
Cross-modal prompt called MaPLe \cite{khattak2023maple} and cross-modal adapter \cite{jiang2022cross} are proposed in both vision and language branches and further achieve cross-modal interactions. 
In CSKT \cite{liu2024clip}, PETL is first successfully incorporated in CLIP for TPR, which provides an effective solution by reducing the number of fine-tuning parameters and training time while achieving comparable performance to full fine-tuning.
In this work, our aim is to develop a more effective and parameter-efficient unified PETL method based on CLIP for TPR, which can transfer more comprehensive information to achieve better performance while fine-tuning fewer parameters.
We focus on three most representative PETL approaches, including Adapter, LoRA and Prefix.

\subsection{Text-based Person Retrieval.}

Text-based Person Retrieval (TPR) was first proposed by Li \emph{et al.} \cite{Li_2017_CVPR} to solve the problem that the target query images are not always available in real-world scenario, \blue{which is a trending topic in intelligent surveillance research \cite{zhai2024zero,zhai2024region,chen2024object,guo2024multiscale,chen2024multi}, which also includes crowd counting, object detection and tracking, person re-identification, anomaly detection, \emph{etc}.}
The central challenge of TPR lies in aligning the person image and text from different modals efficiently.

An early trend in TPR is adopting different uni-modal backbones \cite{shu2022see,chen2022tipcb,farooq2022axm,wu2021lapscore,gao2021contextual,zhu2024improving} such as ResNet, ViT, LSTM or BERT to extract vision and language features, and then two types of representations are aligned by global or local matching methods.
\textbf{Global} matching methods \cite{Li_2017_CVPR,zhang2018deep,chen2022tipcb} align images and texts into a joint embedding space by designing cross-modal matching loss functions.
Although global matching is simple and efficient, it struggles to comprehend more localized information. 
This limitation often leads to poor retrieval performance, even when a synthetic loss function is employed.
Therefore, \textbf{local} matching is proposed to explicitly explore visual-textual salient part pairs for semantic alignment, \emph{e.g.}, human body parts, person strips or regions for image \cite{ding2021semantically,wang2021text,gao2021contextual,shen2023pedestrian}, phrases or words for text \cite{aggarwal2020text,shu2022see,li2021transformer,shen2023pedestrian}. 
Afterwards, \textbf{implicit} matching mechanisms are adopted to extract subtle visual-textual cues such as hairstyle and logo by implicit modeling such as masked language and vision modeling loss \cite{shu2022see,jiang2023cross,farooq2022axm,lin2024cross}.
However, above local and implicit mechanisms typically introduce additional complex multi-level feature extraction or alignment modules. 
Meanwhile, the generalization ability of uni-modal models is limited by their independent single-branch pre-training structure.

Recently, Vision-Language Pre-training (VLP) models \cite{clip,li2021align,li2023blip,li2024adaptive,liu2024causality}  such as CLIP, ALBEF or BLIP have focused on cross-modal representation learning for two modalities simultaneously, which provides more powerful cross-modal capacities from a huge amount of image-text pairs.
Moreover, due to the lightweight architecture of CLIP compared with other more complex VLP models and its abundant image-text corpus, a current trend in text-based person retrieval is to transfer the general knowledge of CLIP to the person domain. 
Specifically, IRRA \cite{jiang2023cross} proposes a cross-modal implicit relation reasoning and aligning framework to achieve CLIP-based fine-grained implicit matching with MLM loss. 
CFine \cite{yan2023clip} designs multi-grained, cross-grained and fine-grained interactions based on CLIP to mine cross-modal correspondences from coarse to fine.
CSKT \cite{liu2024clip} first proposes a novel parameter-efficient method to achieve transfer learning efficiently and effectively under PETL, which introduces a new PETL-based paradigm on CLIP, only training 12M parameters while outperforming the performance of full-tuning CLIP.
In CSKT, two PETL-related methods are constructed: bidirectional prompts on the input side and dual adapters on the output side of Multi-Head Attention (MHA), which provide a collaborative mechanism to prompt cross-modal fusion deeply and transfer knowledge of CLIP with less computational costs.
In this work, we further mine the knowledge within CLIP and aim to propose a more unified PETL framework for TPR. 
\blue{Unlike existing methods which focus on a specific PETL or fully fine-tuning techniques, UP-Person unifies and optimizes multiple PETL methods to leverage their complementary strengths, and designs a robust and easily scalable architecture for TPR. 
It outperforms previous approaches while ensuring parameter efficiency.}

\section{Preliminaries}
\label{sec:Preliminaries}
PETL keeps the pre-trained model frozen and tunes a small number of learnable parameters.
Several state-of-the-art PETL methods are introduced in the following.

\textbf{Adapter.} Adapter \cite{houlsby2019parameter} inserts small modules into transformer layers, which typically employs a down-projection ${\boldsymbol{W}_{{\rm{down }}}} \in {{\mathbb R}^{d \times m}}$ to map the input $x$ to a lower-dimensional space defined by the bottleneck dimension $m$, followed by a nonlinear activation function $f$ like ReLU function and an up-projection with ${\boldsymbol{W}_{{\rm{up}}}} \in {{\mathbb R}^{m \times d}}$.
Finally, adapter is incorporated with a residual connection, the output $x'$ is formulated as:
\begin{eqnarray}
x' \leftarrow x+f\left(x \boldsymbol{W}_{\text {down }}\right) \boldsymbol{W}_{\text {up }}.
\end{eqnarray}
The vanilla sequential structure \cite{houlsby2019parameter} positions two above adapters in series within a layer of the transformer: one following the MHA sublayer and another following the MLP sublayer.
He \emph{et al.} \cite{he2021towards} have proposed an alternative adapter variant that is parallel with MHA or MLP sublayer:
\begin{eqnarray}
h' \leftarrow h\left(x\right)+f\left(x \boldsymbol{W}_{\text {down }}\right) \boldsymbol{W}_{\text {up }},
\end{eqnarray}
where $h(x)$ is the output of the original $x$ by MHA or MLP, and $h'$ represents the final output with a parallel adapter.

\textbf{LoRA.} LoRA    \cite{hu2021LoRA   } incorporates low-rank trainable matrices into transformer layers with the aim of providing an approximation to weight updates. 
For a pre-trained weight matrix $W \in \mathbb{R}^{d \times k}$, LoRA    can update it with a low-rank decomposition:
\begin{eqnarray} 
W+\Delta W=W+\boldsymbol{W}_{\mathrm{down}} \boldsymbol{W}_{\mathrm{up}},
\end{eqnarray}
where $\boldsymbol{W}_{\text {down }} \in \mathbb{R}^{d \times r}$ and $\boldsymbol{W}_{\text {up }} \in \mathbb{R}^{r \times k}$ are learnable. 

For the input $x$ to a linear projection $y=xW$, LoRA  alters the output $y=xW$ to $y'$ in the following manner:
\begin{eqnarray}
y' \leftarrow  x\left(W + s \cdot\boldsymbol{W}_{\text {down }} \boldsymbol{W}_{\text {up }}\right) = xW+s \cdot x \boldsymbol{W}_{\text {down }} \boldsymbol{W}_{\text {up }},
\end{eqnarray} 
where $s$ is a learnable scalar hyper-parameter.
It is noteworthy that LoRA   as a reparameterized method, can merge the weight updates into original weights during inference stage, which decreases computation costs.

\textbf{Prefix and Prompt Tuning.} They incorporate tunable tokens, where the former is added to the input of a transformer block, the latter is prepended to the keys and values of attention.
We will separately explicate the two methods.

Formally, for the input $x \in \mathbb{R}^{n \times d}$ of attention module, the original query, key, and value are denoted as $Q=x{W_{q{\rm{ }}}}$, $K=x{W_{k{\rm{ }}}}$, $V=x{W_{v{\rm{ }}}}$.
Attention (Attn) is formulated as:
\[{\rm{Attn}}\left( {Q,K,V} \right) = {\rm{softmax}}\left( {\frac{{Q{K^T}}}{{\sqrt d }}} \right)V.\]

\textbf{Prefix Tuning (Pre-T)} \cite{li2021prefix} prepends two prefix tokens $\boldsymbol{P}_{k}, \boldsymbol{P}_{v} \in \mathbb{R}^{l \times d}$ to $K,V \in \mathbb{R}^{n \times d}$ respectively.
Therefore, attention is modified as: 
\begin{eqnarray}
y' = {\rm{Attn}}\left( {Q,\left[ {{\boldsymbol{P}_k};K} \right],\left[ {{\boldsymbol{P}_v};V} \right]} \right),
\end{eqnarray}
where $\left[ { \cdot ; \cdot } \right]$ stands for concatenation operation.
The multi-head attention (MHA) with $h$ heads is omitted for brevity.

\textbf{Prompt Tuning (Pro-T)} \cite{lester2021power} prepends prompts $\boldsymbol{P} \in \mathbb{R}^{l \times d} $ to the input tokens $x \in \mathbb{R}^{n \times d}$, which is equivalent to concatenate the same prompt $\boldsymbol{P}W$ to $xW$, denoted as:
\begin{eqnarray}
y' = {\rm{Attn}}([\boldsymbol{P}{W_q};Q],[\boldsymbol{P}{W_k};K],[\boldsymbol{P}{W_v};V]).
\end{eqnarray}

\section{Method}

\begin{figure*}[!ht]
  \centerline{\includegraphics[scale=0.55]{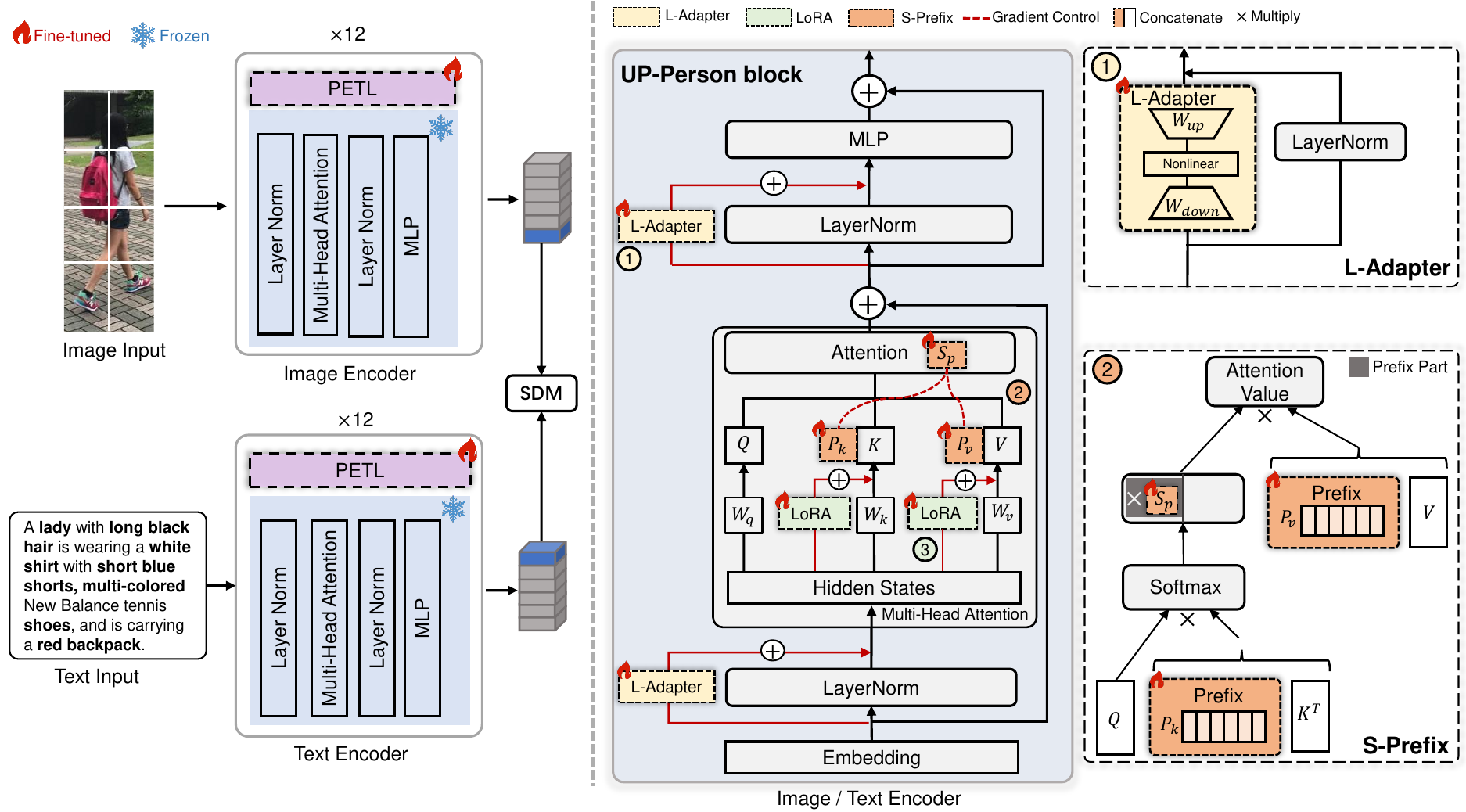}}
  \vspace{-2mm}
  \caption{\textbf{Overview of the proposed UP-Person framework.}
  \textbf{\emph{Left}} is the overall backbone of UP-Person, which consists of image encoder and text encoder based on CLIP, two PETL modules for both encoders, and one parameter-free loss function constraint SDM as optimization objective. Only a few parameters in PETL modules are \emph{fine-tuned} in training phase, while the other original full backbone of CLIP is \emph{frozen}. 
 \textbf{\emph{Right}} is the implementation details of one transformer block for both image and text encoders. In addition to prefix tokens in the keys and values of MHA, S-Prefix proposes a $S_{p}$ factor in attention calculator to enhance gradient propagation of prefix tokens. L-Adapter is proposed in two normalization layers to adjust the overall distribution and avoid submodule conflicts. LoRA is inserted to update the weights of keys and values. Overall, L-Adapter helps transfer global pedestrian features, while LoRA and S-Prefix, working together in MHA, focus on attention to promote local knowledge transferring for TPR. All blocks with dashed borderlines represent the \emph{fine-tuned} modules. On the far right are the more specific implementation details of our L-Adapter and S-Prefix.}
  \label{fig:framework}
\vspace{-3mm}
\end{figure*}

In this section, we illustrate each component of our method in detail.
First, we provide a detailed description of our image and text encoder in Section \ref{ssec:Feature Extraction}.
Then, the architecture of the \textbf{unified PETL} is explained in Section \ref{ssec:Unified}. 
Finally, we present the proposed submodules \textbf{S-Prefix} and \textbf{L-Adapter} in Section \ref{ssec:prefix}.

\subsection{Feature Extraction}
\label{ssec:Feature Extraction}
For a set of person images $I = \{ {I_1},{I_2}, \ldots ,{I_n}\} $ paired with corresponding text queries $T = \{ {T_1},{T_2}, \ldots ,{T_n}\} $, text-based person image retrieval is carried out by evaluating the similarity between each text query and every image, and then the target person image is returned with the highest similarity score.
The initial phase involves feature extraction for both vision and language branches. 

\textbf{Image Encoder.}
We adopt the visual backbone of CLIP (ViT-B/16) \cite{radford2021learning} as our image encoder.
We first partition the image $I \in {\mathbb{R}^{H \times W \times C}}$ into a sequence of $N = H \times W/{P^2}$ non-overlapping patches, where $P$ is the patch size. 
The patches are then mapped to embeddings with a linear projection and added with positional embeddings to enhance spatial information. 
Subsequently, a [CLS] token is introduced at the beginning of the embeddings to denote the overall global information of the image. 
The sequence of $P^2+1$ tokens is then fed into a series of transformer blocks to capture the correlations within these patches.
A transformer block typically consists of a MHA and a MLP, where layer normalization \cite{ba2016layer} is omitted for simplicity in the later formulation.
The input image features are represented as ${X_{i,j,l}} \in {\mathbb{R}^{\left( {{P^2} + 1} \right) \times D}}$, where $l$ is the layer index, $D$ is the hidden dimension of a patch, and $P^2+1$ is the length of the sequence embeddings.
The calculation of $l$-th layer is formulated as:
\begin{eqnarray}
{\hat X_{i,j,l}} =  {\rm MHA}\left( {{X_{i,j,l-1}}} \right) + {X_{i,j,l-1}} \\
{X_{i,j,l}} = {\rm MLP}\left( {{{\hat X}_{i,j,l}}} \right) + {\hat X_{i,j,l}}
\end{eqnarray}

\textbf{Text Encoder.}
The language backbone of CLIP is utilized as text encoder, which is also a 12-layer transformer. 
The computation within a single transformer block mirrors that of the image encoder.
For the input text $T$, 
we tokenize the input description to embeddings $f$ by a simple tokenizer with a 49152 vocab size \cite{sennrich2015neural}.
\blue{For data augmentation, we randomly mask 15\% of the tokens and replace them with the [MASK] token following BERT.}
$f$ then adds [BOS] as the start of the sequence and [EOS] as the end flag.
Thus, the overall sequence can be denoted as $\left\{ {{f_{bos}},{f_1}, \ldots ,{f_{eos}}} \right\}$ and then fed into the transformer as above image encoder by masked MHA,
where the output of $f_{eos}$ is the global representation in language branch.

\subsection{Unified PETL Architecture}
\label{ssec:Unified}
\textbf{Motivation.}
As described in Section \ref{sec:Preliminaries}, we can conclude that different PETL methods generally focus on different parts in the transformer block. 
Specifically, Adapter \cite{gao2023clip} is inserted through a residual connection to adapt the \textbf{output information of MLP and MHA}.
LoRA   \cite{hu2021LoRA   } incorporates low-rank matrices to update weights, which typically represent \textbf{inherent characteristics of a model} for a specific downstream task.
Prefix Tuning \cite{li2021prefix} operates at the forefront of the attention module, which guides the model to focus on more relevant parts of the person image or text description by learnable prefix tokens, thereby extracting more useful information from the \textbf{input of each layer}.
Intuitively, the functions and processed information of the three most representative PETL modules should complement with each other if we carry on careful and proper module design.
Therefore, we infer that a unified PETL framework can exhibit a more powerful expressive capability in text-based person retrieval.

\textbf{Unified PETL Framework.}
Inspired by the above observation and analysis, as shown in Figure \ref{fig:framework}, we propose \textbf{UP-Person} framework based on CLIP backbone, which designs and optimizes multiple lightweight PETL methods: Prefix, LoRA  and Adapter within image and text encoders.
(1) Prefix is optimized to Salable Prefix (S-Prefix) to enhance the adaptation ability of the prefix embeddings for text-based person retrieval task. 
S-Prefix concatenates key and value in attention of MHA sublayer, which complements the task-specific information  and steers model to focus on crucial TPR-specific content.
(2) Low-rank matrices are designed in weight modules of key and value related to MHA to learn more inherently local information about TPR.
(3) Layernorm Adapter (L-Adapter) is proposed to adjust the overall distribution of feature representations in parallel with Layer Normalization (LN), which can adapt shift and bias in a nonlinear way.

\blue{\textbf{Theoretical Analysis of UP-Person.}}
We then provide a more in-depth theoretical justification to \blue{explain the function of each module and why the combination of Prefix, LoRA, and L-Adapter works effectively together in TPR task}.

First, we analyze LoRA component for TPR:
\begin{small}
\begin{align}
Q{{K'}^T}V' &= Q\left( {{K^T} + \Delta {K^T}} \right)\left( {{V^T} + \Delta {V^T}} \right) \nonumber \\
&= \underbrace {Q{K^T}V}_{\text{vanilla attention}} + \underbrace {Q{K^T}\Delta V + Q\Delta {K^T}{V^T} + Q\Delta {K^T}\Delta {V^T}}_{\text{\underline{local information} of TPR}},
\end{align}
\end{small}
where $\Delta K = X\Delta {W_k} = X(\boldsymbol{W}_{down}^k\boldsymbol{W}_{up}^k)$, and $\Delta V = X\Delta {W_v} = X(\boldsymbol{W}_{down}^v\boldsymbol{W}_{up}^v)$.
The second additional term that modifies the attention mechanism enables the model to \blue{capture more nuanced and local features, and relationships in TPR} that the original weight matrices of CLIP cannot fully represent.

Next, the embeddings of Prefix  $P_k$ and $P_v$ can be concatenated on $K$ and $V$ to \blue{injecting task-specific information prompts for TPR, which helps the attention mechanism focus on task-relevant information}.
The attention with Prefix can be expressed as:

\begin{equation}
\small
\begin{array}{l}
Q{\left[ {{P_k};K'} \right]^T}\left[ {{P_v};V'} \right] = Q{{K'}^T}V' + Q{P_k}^T{P_v}\\
= \underbrace {Q{K^T}V}_{\text{vanilla attention}} + \underbrace {Q{K^T}\nabla V + Q\nabla {K^T}{V^T} + Q\nabla {K^T}\nabla {V^T}}_{\text{\underline{local information} of TPR}} + \\ \underbrace {Q{P_k}^T{P_v}.}_{\text{\underline{task-specific information} of TPR}}
\end{array}
\end{equation}

Finally, Adapter component can be utilized to adjust the overall distribution in layernorm, fine-tuning intermediate features from a global perspective:
\begin{eqnarray}
y \to y + \underbrace {f(y \cdot {W_{down}}){W_{up}},}_{\text{\underline{global information} of TPR}}
\end{eqnarray}
where $y={LN(Q{{\left[ {{P_k};K'} \right]}^T}\left[ {{P_v};V'} \right]})$, and $y$ represents the normalized intermediate features.

\blue{The above PETL submodules do not conflict in terms of functionality or spatial distribution. Theoretically, the different components work synergistically to enhance performance.}

\blue{Therefore, by introducing these additional components to fine-tune the original features from coarse to fine, we can cohesively enhance global distribution, local dependencies, and  TPR-specific prompt information.}

we further introduce two our novel submodules: \textbf{S-Prefix} and \textbf{L-Adapter}.
\subsection{S-Prefix} 
\label{ssec:prefix}

Compared with the prompt-based submodule in CSKT \cite{liu2024clip}, 
Prefix is more flexible and efficient when attached to multiple layers, since it does not change sequence length.
According to the study \cite{he2021towards,gao2023unified}, Prefix can be represented as a form similar to adapter, which can be viewed as working on the original head attention output $h$:
\begin{eqnarray}
h \leftarrow (1 - \lambda (x))h + \lambda (x){\mathop{\rm softmax}\nolimits} \left( {x{W_q}\boldsymbol{P_k}^{\rm T}} \right)\boldsymbol{P_v},
\label{eq:prefix}
\end{eqnarray}
where $\lambda (x)$ is formulated as:

\begin{eqnarray}
\lambda (x) = \frac{{\sum\limits_i {\exp } {{\left( {x{W_q}\boldsymbol{P_k}^ \top } \right)}_i}}}{{\sum\limits_i {\exp } {{\left( {x{W_q}\boldsymbol{P_k}^ \top } \right)}_i} + \sum\limits_j {\exp } {{\left( {x{W_q}W_k^ \top {x^ \top }} \right)}_j}}}.
\end{eqnarray}

However, during the training phase, it was observed that the convergence rate of the vanilla prefix was notably slow in TPR task, resulting in a poor retrieval performance.
Our analysis indicates that this phenomenon is mainly caused by relatively small gradient values of prefix tokens.
Specifically, the gradient of $\boldsymbol{P_v}$ \cite{gao2023unified} can be denoted as:
\begin{eqnarray}
\frac{{\partial {\cal L}}}{{\partial {\boldsymbol{P_v}}}} = {\left( {\frac{{\partial h}}{{\partial {\boldsymbol{P_v}}}}} \right)^ \top }\frac{{\partial {\cal L}}}{{\partial h}} = \lambda (x){\left( {\sigma \left( {x{W_q}{\boldsymbol{P_k}}^ \top } \right)} \right)^ \top }\frac{{\partial {\cal L}}}{{\partial h}},
\end{eqnarray}
where $\sigma$ is the softmax function. 
Since the length of $\boldsymbol{P_k}$ is much less than input $x$, $\lambda (x)$ is a small value tending to 0, which dramatically reduces the convergence speed of training, and further significantly impacts the retrieval performance of TPR task.
A similar conclusion can be drawn on $\boldsymbol{P_k}$.

\begin{figure}[!tb]
  \centerline{\includegraphics[width=0.44\textwidth,height=0.43\textwidth]{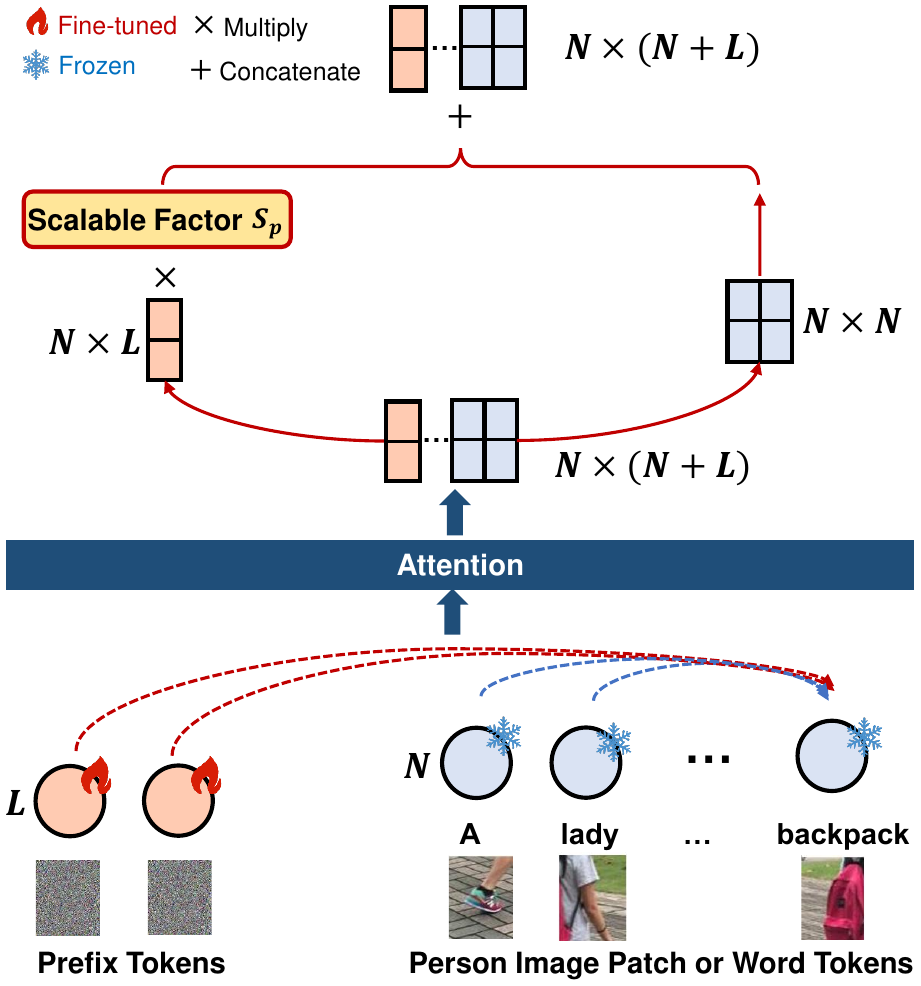}}
  \vspace{-2mm}
  \caption{\textbf{Illustration of S-Prefix.} We utilize ${S_{p}}$ to denote the salable factor about attention of prefix to accelerate the convergence rate. S-Prefix submodules are inserted in all transformer layers of two branches.}
  \label{fig:prefix}
\vspace{-3mm}
\end{figure}

To solve this issue,  we propose an improved module  \textbf{Salable Prefix (S-Prefix)} to enhance gradient propagation of prefix tokens, which introduces a novel salable factor on prefix-related attention to optimize original prefix.
As shown in Figure \ref{fig:prefix}, we first convert the $N$ input tokens $X \in \mathbb{R}^{N \times D}$ into queries $Q$, keys $K$ and values $V$.
Then, the $L$ prefix tokens are inserted into $K$ and $V$, and the transformed keys and values are denoted as
$K^{'}, V^{'} \in \mathbb{R}^{(N+L) \times D}$.
After applying the attention mechanism, $QK^{'T} \in \mathbb{R}^{N\times(N+L)}$. 
S-Prefix separates attention matrix into two parts: prefix matrix and original matrix.
we design a salable factor $\boldsymbol{S_{p}}$ to boost the attention related to the prefix part and further speed up the convergence, which is a lightweight multiplier for the attention module of prefix.
In order to demonstrate how and where ${S_{p}}$ works effectively, S-Prefix is equivalent to inserting ${S_{p}}$ to Equation (\ref{eq:prefix}), finally denoted as:
\begin{eqnarray}
h \leftarrow (1 - \lambda (x))h + \boldsymbol{S_p} \cdot \lambda (x)    \cdot {\mathop{\rm softmax}\nolimits} \left( {x{W_q}\boldsymbol{P_k}^{\rm T}} \right)\boldsymbol{P_v}.
\end{eqnarray}
Subsequently, the improved gradient of prefix is as follows:
\begin{eqnarray}
\frac{{\partial {\cal L}}}{{\partial {\boldsymbol{P_v}}}} = {\left( {\frac{{\partial h}}{{\partial {\boldsymbol{P_v}}}}} \right)^ \top }\frac{{\partial {\cal L}}}{{\partial h}} =  \boldsymbol{S_p} \cdot \lambda (x){\left( {\sigma \left( {x{W_q}{\boldsymbol{P_k}}^ \top } \right)} \right)^ \top }\frac{{\partial {\cal L}}}{{\partial h}},
\end{eqnarray}
\blue{where $\boldsymbol{S_{p}}$ is a learnable factor designed to address the issue of excessively small gradients for $\boldsymbol{P_v}$ caused by $\lambda(x)$.}

\subsection{L-Adapter.}
\begin{figure*}[!tb]
\centerline{\includegraphics[width=0.91\textwidth,height=0.32\textwidth]{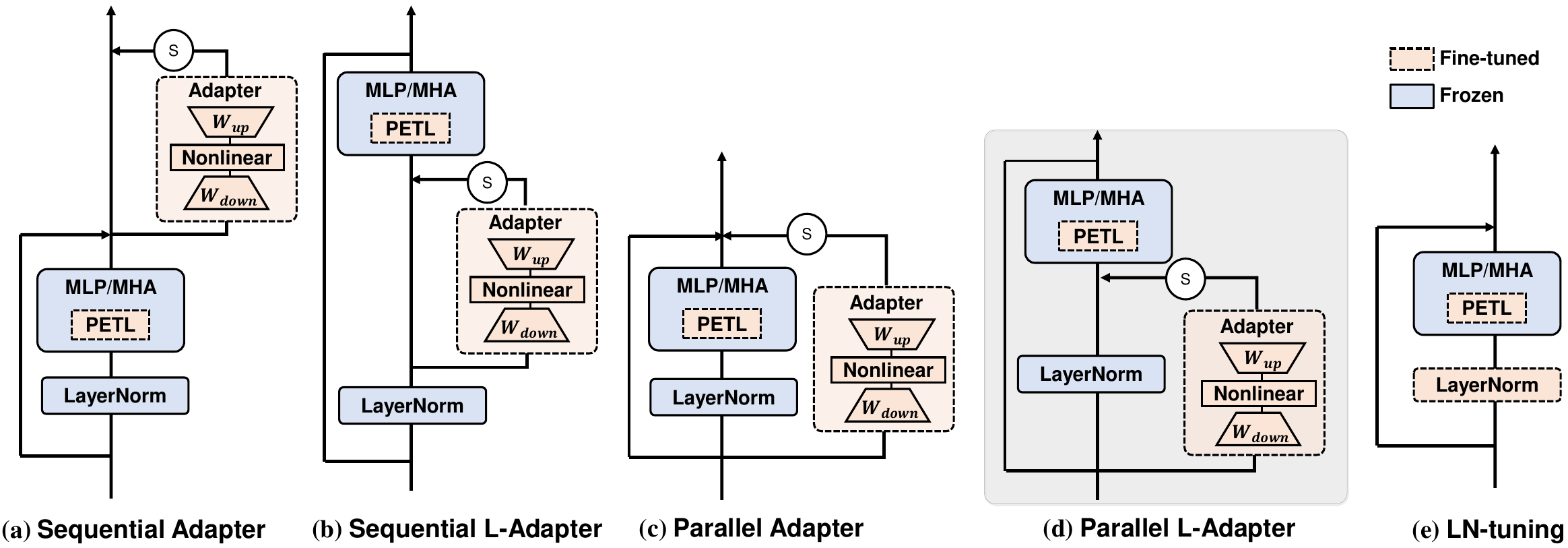}}
\vspace{-2mm}
  \caption{\textbf{Illustration of L-Adapter.} (a) Sequential Adapter is connected behind MLP or MHA. (b) Sequential L-Adapter is connected behind layernorm. (c) Parallel Adapter always spans layernorm and MLP or MHA which contains other PETL submodules. (d) Parallel L-Adapter is inserted into layernorm with residual connection, which is separated from other PETL submodules and transfers knowledge independently. (e) LN-tuning unfreezes layernorm, making the original features fine-tuned directly. }
  \label{fig:adapter}
\vspace{-3mm}
\end{figure*}

The vanilla adapters \cite{houlsby2019parameter,chen2022adaptformer} typically act on modules containing MLP or MHA, either in a sequential or parallel manner as Figure \ref{fig:adapter}(a)(c). 
However, in the unified framework, PETL submodules such as S-Prefix and LoRA are designed \emph{within} MHA or MLP for functional diversity on different locations. 
The original parallel adapter and other PETL submodules simultaneously do fine-tuning on input as Figure \ref{fig:adapter}(c), and their structures overlap and interact, which is prone to disrupting the intended optimization direction and causes component conflicts.
Unlike them, in this paper, we focus on Layer Normalization to get rid of causing conflicts from space and explore appropriate connection manner including sequential and parallel.

Layer Normalization (LN) \cite{ba2016layer}, also known as layernorm, is used to standardize the distributions of intermediate layers. 
This normalization process enhances the smoothness of gradients during training, accelerates the training process, and improves generalization accuracy.
The procedure unfolds in two main steps:
(1) Normalization of $x$ by mean and variance.
This helps in centering and scaling the values, bringing $x$ to a standard form.
(2) Scaling and shifting operations with gain $g$ and bias $b$.
Following the normalization step, the standardized values undergo a scaling and shifting process. 
The scaling operation, governed by the gain term $g$, allows for adjusting the spread of the values. 
The shifting operation, regulated by the bias term $b$, enables the network to introduce necessary variations.
In essence, layernorm not only ensures that intermediate layers have consistent statistical properties but also introduces adaptability through the scaling and shifting operations.
we can conclude that layernorm is important in fine-tuning to downstream tasks, and unfreezing layernorm (LN-tuning) like \cite{qi2022parameter} has been explored.
However, LN-tuning as Figure \ref{fig:adapter}(e) directly fine-tunes the shift and scale parapmeters, which damages the inherent features of CLIP in a linear way.

Inspired by the above, we believe that further exploring the powerful transferring ability of layernorm is promising. 
As depicted in Figure \ref{fig:adapter}(d), \textbf{Layernorm Adapter (L-Adapter)} is finally designed in parallel with layernorm to \emph{adjust overall distribution in a nonlinear way}, which is more flexible and capable of addressing more complex situations compared to linear methods such as LN-tuning, and can effectively \emph{avoid module conflicts} compared with the vanilla parallel adapter.
It can be formulated as:
\begin{eqnarray} 
h \leftarrow {\rm LayerNorm}\left( x \right) + s \cdot {\rm Adapter} \left( x \right).
\end{eqnarray}

\subsection{Optimization Objective} 
\label{ssec:train and infer}
A parameter-free loss function is adopted in training phase termed as Similarity Distribution Matching (SDM) \cite{jiang2023cross}, which integrates the cosine similarity distributions of the $N\times N$ embeddings for image-text pairs into the KL divergence to build up the connection of two modalities.

For a mini-batch containing $N$ image-text pairs, we form a set of image-text representation pairs as $\left\{\left(f_i^v, f_j^t\right), y_{i,j}\right\}_{j=1}^N$, where $y_{i,j} =1$ represents a matched pair from the same person, and $y_{i,j} =0$ indicates an unmatched pair.
The probability of a matching pair $p_{i, j}$ is calculated with a softmax function as follows:
\begin{eqnarray} 
p_{i, j}=\frac{\exp \left(sim\left(f_i^v, f_j^t\right) / \tau\right)}{\sum_{k=1}^N \exp \left(sim\left(f_i^v, f_k^t\right) / \tau\right)},
\end{eqnarray}
where $sim(f^v, f^t)$ denotes cosine similarity between text embedding $f^v$ and image embedding $f^t$, and $\tau$ is a temperature hyper-parameter that controls the sharpness of the probability distribution.
We denote $q_{i, j}=y_{i, j} / \sum_{k=1}^N y_{i, k}$ as the true matching probability.
Finally, the SDM loss from image to text is computed based on above probabilities and KL divergence:
\begin{eqnarray} 
\mathcal{L}_{i 2 t}=K L\left(\mathbf{p}_{\mathbf{i}} \| \mathbf{q}_{\mathbf{i}}\right)=\frac{1}{N} \sum_{i=1}^N \sum_{j=1}^N p_{i, j} \log \left(\frac{p_{i, j}}{q_{i, j}+\epsilon}\right),
\label{eq:i2t}
\end{eqnarray}
where $\epsilon$ is a small number to avoid potential issues with numerical calculations.
Then, the bidirectional SDM loss is formulated as:
\begin{eqnarray} 
\mathcal{L}_{s d m}=\mathcal{L}_{i 2 t}+\mathcal{L}_{t 2 i},
\end{eqnarray}
\blue{where $\mathcal{L}_{i 2 t}$ denotes image-to-text matching for the input text, and $\mathcal{L}_{t 2 i}$ denotes text-to-image matching for the input image, similar to Equation (\ref{eq:i2t}). 
Both weights of the bidirectional loss functions are set to 1 equally  to enhance cross-modal alignment ability.}

\section{Experimental Settings}
\label{sec:expe}
This section introduces the three official datasets, implementation details and evaluation metrics.

\textbf{CUHK-PEDES} as the most commonly used dataset, contains 40,206 images and 80,412 textual descriptions for 13,003 identities. The training set consists of 11,003 identities with 34, 054 images and 68, 126 texts.
Both the validation set and test set have 1,000 identities, where the former contains 3, 078 images and 6, 158 texts, and the latter has 3, 074 images and 6, 156 texts.
\textbf{ICFG-PEDES} contains 54,522 images for 4,102 identities. Each image corresponds to one description. The training and test sets contain 3,102 identities with 34, 674 images, and 1,000 identities with 19, 848 identities, respectively.
\textbf{RSTPReid} as a newly released dataset contains 20,505 images of 4,101 identities. Each image has 2 descriptions. The training, validation and test sets contain 3701 identities with 18505 images, 200 identities with 1000 images, and 200 identities with 1000 images respectively.

\textbf{Implementation Details.}
UP-Person consists of a pre-trained image encoder, \emph{i.e.}, CLIP-ViT-B/16, a pre-trained text encoder, \emph{i.e.}, CLIP text Transformer, and PETL modules including S-Prefix, LoRA and L-Adapter.
The image is resized to 384 × 128, and the length of textual token sequence is 77.
UP-Person is trained using Adam optimizer for 60 epochs, with a batch size of 128 and an initial learning rate $1 \times {10^{ - 3}}$ in CUHK-PEDES and ICFG-PEDES.
As the PETL methods are sensitive to learning rate with the change of the data scale.
we separately set an initial learning rate $1 \times {10^{ - 4}}$ in RSTPReid due to the small data scale.
\blue{Each block of CLIP, spanning 12 layers, incorporates S-Prefix, L-Adapter, and LoRA components as illustrated in our framework.
Specifically, S-Prefix and LoRA are integrated into MHA.
Meanwhile, L-Adapter spans in parallel within LayerNorm of both  MHA and MLP.
The bottleneck dimension $b$ of the L-Adapter is set to 8 across all three datasets. 
For CUHK-PEDES and ICFG-PEDES, the rank of LoRA $r$ is set to 32, while for RSTPReid, due to its smaller dataset size, rank is set to 16 for better performance. 
Regarding the length of S-Prefix $l$, the values are set to 10 for CUHK-PEDES, 14 for ICFG-PEDES, and 2 for RSTPReid.}
The salable factor for $S_p$ is initialized to 10.
We set the temperature parameter $\tau$ in SDM to 0.02.
We perform our experiments on a single NVIDIA 4090 24GB GPU.

\textbf{Evaluation Metrics.} Rank-k metrics (k=1,5,10) are adopted as the primary evaluation metrics, which denote the probability of finding at least one matching person image within the top-k candidates when given a textual description. 
Additionally, we adopt the mean Average Precision (mAP) as a comprehensive retrieval criterion. 
The higher Rank-k, mAP indicates better performance.

\section{Experimental Results and analysis}

\subsection{Performance Comparison}
We compare our UP-Person with other methods on three datasets: CUHK-PEDES, ICFG-PEDES and RSTPReid.
The experimental results are shown in Table \ref{tab:cuhk}, \ref{tab:icfg} and \ref{tab:rstp} respectively.
UP-Person outperforms other fully fine-tuned CLIP-based methods such as IRRA-CLIP, IRRA and CFine with R@1 achieving 74.17\%, 65.02\% and 63.15\% on three datasets respectively while only fine-tuning a few parameters.
Moreover, we demonstrate that full-tuning method IRRA has a high risk of overfitting, which is particularly evident when the training dataset is smaller, referring to Table \ref{tab:cuhk} (larger dataset CUHK-PEDES) and Table \ref{tab:rstp} (smaller dataset RSTPReid). 
UP-Person on RSTPReid outperforms IRRA by a larger margin \emph{i.e.}, +2.95\% on R@1.
Meanwhile, compared with the current best-performing PETL-based method DM-Adapter, UP-Person achieves a significant improvement in terms of all metrics.
It outperforms DM-Adapter by 2.00\%, 2.38\% and 3.15\% on R@1 across three datasets with less fine-tuned parameters.
Furthermore, with the CLIP-Large model (CLIP-ViT L/14), the performance has been boosted to 76.04\% on CUHK-PEDES.
UP-Person (ViT-14/L) has shown comparable performance to the SOTA method APTM (with large-scale generated dataset MALS). 
In the following comparisons, we adopt UP-Person with the ViT B/16 model, ensuring a consistent evaluation with the other CLIP-based baselines.

\begin{figure}[!tb]
  \centerline{\includegraphics[scale=0.47]{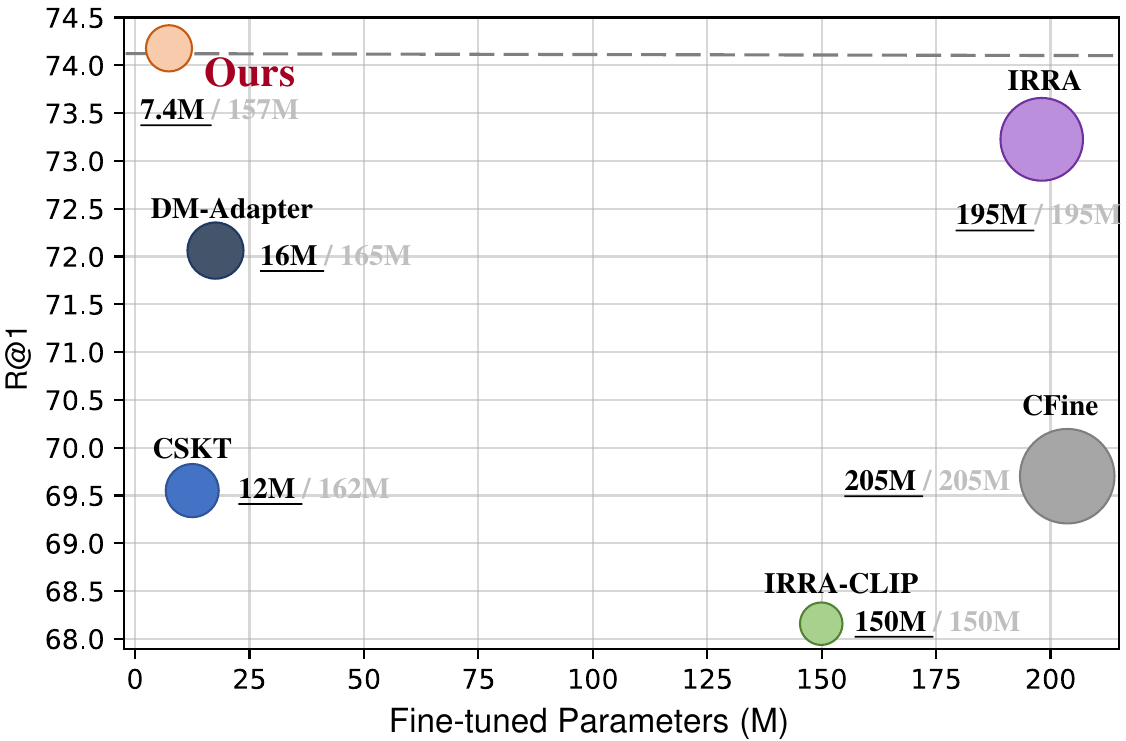}}
  \vspace{-2mm}
  \caption{R@1 and parameters of different CLIP-based methods on CUHK-PEDES.The horizontal coordinate denotes the number of fine-tuned parameters. \blue{The gray numbers and the radius of the circles both represent the entire model size.}}
  \label{fig:compare}
\vspace{-3mm}
\end{figure}

Figure \ref{fig:compare} depicts the trade-off between fine-tuned parameters and retrieval performance on CUHK-PEDES among CLIP-based methods, where UP-Person achieves optimal performance with the minimum number of parameters.
Our approach only stores 7.4M extra parameters in one scenario under the condition of sharing a frozen CLIP model, which greatly improves storage efficiency when having multiple scenarios (multiple datasets).

Table \ref{tab:efficiency} provides a more holistic comparison of our UP-Person with other CLIP-based methods in efficiency and effectiveness. 
UP-Person utilizes only 3262 MB of memory during training, which is less than half of IRRA's memory usage.
CSKT, DM-Adpater and UP-Person are based on PETL and demonstrate greater advantages in memory efficiency. 
In addition to the basic fully fine-tuned CLIP method (IRRA-CLIP), UP-Person has the lowest FLOPs and the number of model parameters, indicating lower computational complexity. 
\blue{For the trade-off between performance and inference latency, in UP-Person, LoRA modifies the inner weights directly and inference without introducing any computational overhead. 
While the inclusion of S-Prefix and L-Adapter components introduce some additional computational load compared to pure CLIP, this trade-off is minor and negligible relative to the substantial performance gains.
As shown in Table \ref{tab:efficiency}, UP-Person demonstrates a clear advantage over all CLIP-based methods in performance.
Furthermore, in terms of inference efficiency, UP-Person is comparable to the other PETL-based methods, and it also far outperforms the multi-grained CFine.
UP-Person ranks among the top two in multiple metrics and achieves a favorable trade-off between performance and inference efficiency.}
Therefore, considering metrics such as computational efficiency, storage efficiency, model complexity, and memory usage, UP-Person exhibits a notable advantage over other methods.

\begin{table*}[tb]
\small
\centering
\tabcolsep=5pt

\renewcommand\arraystretch{1.1}

\caption{Performance Comparison with other methods on CUHK-PEDES. The left column denotes whether using  CLIP.  ``G'' and ``L'' in ``Type'' denote global and local matching. ``P'' stands for the PETL-related methods (such as CSKT and ours). }
\resizebox{0.9\textwidth}{!}{%
\begin{tabular}{c|l|cccccccc}
\hline
                                               & Method  & Type & Ref & Image Enc. & Text Enc. & R@1  & R@5  & R@10   & mAP  \\
\hline
\multirow{9}{*}{\rotatebox{90}{w/o CLIP}}      
                                            & CMPM/C \cite{zhang2018deep}   &  L & ECCV18 & RN50 & LSTM & 49.37     & 71.69      & 79.27      & -        \\
                                               & ViTAA \cite{wang2020vitaa}  &  L & ECCV20 & RN50 & LSTM    & 55.97     & 75.84      & 83.52      & -      \\
                                               & NAFS \cite{gao2021contextual}   & L &arXiv21  & RN50 & BERT & 59.36 & 79.13 & 86.00 & 54.07 \\
                                               & DSSL \cite{zhu2021dssl}   &  L   & MM21 & RN50 & BERT     & 59.98     & 80.41      & 87.56      & -      \\

                                               & SSAN \cite{ding2021semantically} &  L   & arXiv21 & RN50 & LSTM      & 61.37     & 80.15      & 86.73      & -      \\   
                                               & SAF \cite{li2022learning}   &  L &ICASSP22 & ViT-Base & BERT    & 64.13     & 82.62      & 88.40      & 58.61  \\
                                               & TIPCB \cite{chen2022tipcb}  &  L &Neuro22 & RN50 & BERT   & 64.26     & 83.19      & 89.10      & -      \\
                                               & AXM-Net \cite{farooq2022axm}  &  L & AAAI22 & CNN Blocks & BERT \& CNN Blocks    & 64.44     & 80.52      & 86.77      & 58.73  \\
                                               & TGDA  \cite{gao2023addressing}    & L     & TCSVT23 &RN50 & BERT   & 64.64     & 83.38     & 89.34      & 58.64 \\
                                               & LGUR \cite{shao2022learning}  &  L & MM22 & DeiT-Small & BERT & 65.25     & 83.12      & 89.00      & -  \\
                                               & IVT \cite{shu2022see}    & G     & ECCV22 &ViT-Base & BERT   & 65.59     & 83.11      & 89.21      & -  \\
                                              & Wu et al. \cite{wu2023contrastive} & G  & TCSVT23 &ViT-Base & BERT & 69.47 &87.13& 92.13& 60.56 \\
                                             & APTM (w/o MALS) \cite{yang2023towards} & L  & MM23 &Swin Transformer & BERT & 66.44 & 84.92 & 90.76 &59.19 \\
                                             & APTM (w/ MALS) \cite{yang2023towards} & L  & MM23 &Swin Transformer & BERT & 76.53 & 90.04 & 94.15 & 66.91 \\

\hline\hline
\multirow{6}{*}{\rotatebox{90}{w/ CLIP}}        & Han et al. \cite{han2021text} & G  &BMVC21 &CLIP-RN101 & CLIP-Transformer   & 64.08     & 81.73      & 88.19      & 60.08  \\

                                               & CFine \cite{yan2023clip}   & L   &TIP23& CLIP-ViT-Base & BERT    & 69.57     & 85.93      & 91.15      & -  \\
                                               & IRRA-CLIP \cite{jiang2023cross} & G &CVPR23 & CLIP-ViT-Base & CLIP-Transformer & 68.19     &86.47      & 91.47      & 61.12           \\
                                                & IRRA$^*$ \cite{jiang2023cross} & G &CVPR23 & CLIP-ViT-Base & CLIP-Transformer & 71.15 & 87.66 & 92.58 & 64.84 \\
                                                & IRRA  \cite{jiang2023cross} & G &CVPR23 & CLIP-ViT-Base & CLIP-Transformer & 73.38 & 89.93 & 93.71 & 66.13 \\ 
                                                
\noalign{\vskip-1mm}                                                
\cmidrule[1\arrayrulewidth]{2-10}
\noalign{\vskip-1mm}

                                               &\cellcolor{gray!40}CSKT \cite{liu2024clip} (Baseline)  & \cellcolor{gray!40} P+G &\cellcolor{gray!40} ICASSP24   & \cellcolor{gray!40} CLIP-ViT-Base & \cellcolor{gray!40}CLIP-Transformer & \cellcolor{gray!40}69.70    & \cellcolor{gray!40} 86.92    &\cellcolor{gray!40} 91.80    & \cellcolor{gray!40}62.74 \\

                                          &\cellcolor{gray!40}DM-Adapter \cite{liu2025dm}  & \cellcolor{gray!40} P+G &\cellcolor{gray!40} AAAI2025   & \cellcolor{gray!40} CLIP-ViT-Base & \cellcolor{gray!40}CLIP-TE & \cellcolor{gray!40}\underline{72.17}    & \cellcolor{gray!40} \underline{88.74}    &\cellcolor{gray!40} \underline{92.85}    & \cellcolor{gray!40}\underline{64.33} \\

                                               &   \cellcolor{gray!40}\textbf{UP-Person (Ours, ViT-16/B)}   &  \cellcolor{gray!40} \textbf{P+G} &\cellcolor{gray!40} - &  \cellcolor{gray!40} CLIP-ViT-Base &   \cellcolor{gray!40} CLIP-Transformer & \cellcolor{gray!40}\textbf{74.17} & \cellcolor{gray!40} \textbf{89.70} &  \cellcolor{gray!40} \textbf{93.88} & \cellcolor{gray!40}\textbf{65.91} \\

\cline{2-10}

                                               &   \cellcolor{gray!40}\textbf{UP-Person (Ours, ViT-14/L)}   &  \cellcolor{gray!40} \textbf{P+G} &\cellcolor{gray!40} - &  \cellcolor{gray!40} CLIP-ViT-Large &   \cellcolor{gray!40} CLIP-Transformer & \cellcolor{gray!40}76.04 & \cellcolor{gray!40} 90.30 &  \cellcolor{gray!40} 94.80 &\cellcolor{gray!40}68.00 \\                                               


\hline
\end{tabular}
}

\vspace{1mm} 
$*$ indicates our replication results after a minor bug correction, also viewed as data augmentation in vanilla IRRA.\\
\label{table1}
 \label{tab:cuhk}
 \vspace{-3mm}
\end{table*}

\begin{table}[tb]
\small
\centering
\tabcolsep=5pt
\renewcommand\arraystretch{1.1}
\caption{Comparison on ICFG-PEDES.}
\vspace{-2mm}
\resizebox{0.4\textwidth}{!}{%
\begin{tabular}{c|l|cccc}
\hline
                                               & Method    & R@1  & R@5  & R@10   & mAP \\
\hline
\multirow{7}{*}{\rotatebox{90}{w/o CLIP}}                                                                                           & CMPM/C \cite{zhang2018deep}      & 
                                              43.51     & 65.44     & 74.26      & -  \\
                                               & ViTAA \cite{wang2020vitaa}       & 50.98     & 68.79     & 75.78      & -      \\
                                               & SSAN \cite{ding2021semantically} & 54.23     & 72.63     & 79.53      & -      \\
                                               & SAF \cite{li2022learning}        & 54.86     & 72.13     & 79.13      & 32.76    \\
                                               & TIPCB \cite{chen2022tipcb}       & 54.96     & 74.72     & 81.89      & -      \\
                                             & IVT \cite{shu2022see}            & 56.04     & 73.60     & 80.22      & -  \\
                                             & TGDA  \cite{gao2023addressing}   & 57.26     & 75.19      & 81.80      & 32.27  \\
                                             & Wu et al. \cite{wu2023contrastive} &
                                               57.69 &75.79& 82.67 &36.07 \\
                                               & LGUR \cite{shao2022learning}     & 59.02     & 75.32     & 81.56      & -  \\
                                               & APTM (w/o MALS) \cite{yang2023towards} & 57.49 & 75.84 & 82.60 &32.41 \\
                                             & APTM (w/ MALS) \cite{yang2023towards} & 68.51 & 82.99 &87.56& 41.22 \\

\hline\hline
\multirow{5}{*}{\rotatebox{90}{w/ CLIP}}
                                               & CFine \cite{yan2023clip}         & 60.83    & 76.55      & 82.42      & -  \\
                                               & IRRA-CLIP \cite{jiang2023cross} & 56.74     & 75.72      & 82.26      & 31.84           \\
                                               & IRRA$^*$ \cite{jiang2023cross}   & 61.36 & 78.66 & 84.60 & 37.95         \\
                                               & IRRA \cite{jiang2023cross}   &  63.46  & 80.25 &85.82 & 38.06\\ 
                                               & \cellcolor{gray!40}CSKT \cite{liu2024clip}      & \cellcolor{gray!40}58.90    & \cellcolor{gray!40}77.31    &\cellcolor{gray!40}83.56    & \cellcolor{gray!40}33.87 \\
                                            & \cellcolor{gray!40}DM-Adapter \cite{liu2025dm}       & \cellcolor{gray!40}\underline{62.64}    & \cellcolor{gray!40}\underline{79.53}    &\cellcolor{gray!40}\underline{85.32}    & \cellcolor{gray!40}\underline{36.50} \\
                                               
                                               & \cellcolor{gray!40}\textbf{UP-Person (Ours, ViT-16/B)}    & \cellcolor{gray!40}\textbf{65.02}    &\cellcolor{gray!40}\cellcolor{gray!40}\textbf{80.98} & \cellcolor{gray!40}\textbf{86.16} &\cellcolor{gray!40}\textbf{38.32} \\ 
                                               \cline{2-6}
                                    & \cellcolor{gray!40}\textbf{UP-Person (Ours, ViT-14/L)} & \cellcolor{gray!40}65.98 & \cellcolor{gray!40}81.94 & \cellcolor{gray!40}87.05 & \cellcolor{gray!40}40.12 \\

\hline
\end{tabular}}
\label{table2}
\label{tab:icfg}
\vspace{-2mm}
\end{table}

\begin{table}[tb]
\small
\centering
\tabcolsep=5pt
\renewcommand\arraystretch{1.1}
\vspace{-2mm}
\caption{Comparison on RSTPReid.}
\vspace{-2mm}
\resizebox{0.4\textwidth}{!}{%
\begin{tabular}{c|l|cccc}
\hline
                                               & Method    & R@1  & R@5  & R@10   & mAP \\
\hline
\multirow{4}{*}{\rotatebox{90}{w/o CLIP}}       & DSSL \cite{zhu2021dssl}          & 32.43                                                  & 55.08      & 63.19      & -      \\
                                               & SSAN \cite{ding2021semantically} & 43.50     & 67.80      & 77.15      & -      \\
                                               & SAF \cite{li2022learning}        & 44.05     & 67.30      & 76.25      & 36.81    \\
                                               & IVT \cite{shu2022see}            & 46.70     & 70.00     & 78.80      & -  \\
                                              & TGDA  \cite{gao2023addressing}     & 48.35     & 73.15     & 80.30      & 37.96  \\
                                             & APTM (w/o MALS) \cite{yang2023towards} & 47.20 & 70.85 &  80.00 & 36.36 \\
                                             & APTM (w/ MALS) \cite{yang2023towards} & 67.50& 85.70 &91.45 &52.56 \\

\hline\hline
\multirow{5}{*}{\rotatebox{90}{w/ CLIP}}       
                                               & CFine \cite{yan2023clip}         & 50.55     & 72.50     & 81.60      & -  \\
                                               & IRRA-CLIP \cite{jiang2023cross} & 54.05     & 80.70     & 88.00	& 43.41        \\
                                               & IRRA$^*$ \cite{jiang2023cross} & 57.50 & 80.15 & 87.05 & 44.31       \\
                                              & IRRA \cite{jiang2023cross} & 60.20 & 81.30 & 88.20 & 47.17       \\ 
                                               &\cellcolor{gray!40}CSKT \cite{liu2024clip}    & \cellcolor{gray!40}57.75   &\cellcolor{gray!40}81.30   &\cellcolor{gray!40}88.35 & \cellcolor{gray!40}46.43 \\

                                               &\cellcolor{gray!40}DM-Adapter \cite{liu2025dm}    & \cellcolor{gray!40}\underline{60.00}   &\cellcolor{gray!40}\underline{82.10}   &\cellcolor{gray!40}87.90 & \cellcolor{gray!40}\underline{47.37} \\
                                               & \cellcolor{gray!40}\textbf{UP-Person (Ours, ViT-16/B)}   &\cellcolor{gray!40}\textbf{63.15} &\cellcolor{gray!40}\textbf{83.45}
                                               &\cellcolor{gray!40}\textbf{89.75}
                                               &\cellcolor{gray!40}\textbf{48.15} \\ 
                                           &\cellcolor{gray!40}\textbf{UP-Person (Ours, ViT-14/L)}   &\cellcolor{gray!40}64.45 &\cellcolor{gray!40}84.75
                                               &\cellcolor{gray!40}90.95
                                               &\cellcolor{gray!40}51.17 \\

\hline
\end{tabular}}
\label{table3}
\label{tab:rstp}
\vspace{-2mm}
\end{table}

\begin{table*}[!tb]
    \centering
    \caption{Analysis of efficiency and effectiveness. To ensure a fair comparison of memory costs, batch size for all methods in the training phase is set to 32. The inference time is tested on CUHK-PEDES.}
    \renewcommand\arraystretch{1.1} 
    \resizebox{0.95\textwidth}{!}{%
        \begin{tabular}{lcccccccc}
            \hline
            Method  & R@1 CUHK$\uparrow$ & R@1 ICFG $\uparrow$ & R@1 RSTP $\uparrow$ & Memory Cost (M) $\downarrow$  & Trainable \#Param (M) $\downarrow$ & Model \#Param (M) $\downarrow$ & FLOPs $\downarrow$ & Inference Time (s) $\downarrow$ \\
            \hline
            CFine      & 69.57 & 60.83 & 50.55 & 13570 (55.24\%)  & 205M & 205M & 20159.03M  & 28.38\\
            IRRA-CLIP      & 68.19& 56.74 & 54.05 & 4474 (18.21\%)   & 150M  & \textbf{150M} & \textbf{12979.27M} & \textbf{16.21} \\
            IRRA       & \underline{73.38} & \underline{63.46} & \underline{60.20} & 7034 (28.64\%)   & 195M  & 195M & 17542.28M & \textbf{16.21} \\
            CSKT    & 69.70 & 58.90 & 57.75 & \textbf{2338 (9.52\%})  & \underline{12M} & 161M & 13886.23M & 18.10\\
            DM-Adapter   & 72.17 & 62.64 & 60.00 & \underline{2952 (12.02\%})  & 16M & 165M & 13796.91M & 18.79\\
            \textbf{UP-Person (Ours)} & \textbf{74.17} &  \textbf{65.02} &  \textbf{63.15} & 3262 (13.28\%)  & \textbf{7.4M}  & \underline{157M} &  \underline{13783.40M} & 18.17\\ 
            \hline
        \end{tabular}
    }
    \label{tab:efficiency}
    \vspace{-3mm}
\end{table*}


\subsection{Ablation Study}
\textbf{Ablations on PETL components.}
We analyze the effectiveness and contribution of each PETL module and synergetic effects by conducting a series of ablation experiments on CUHK-PEDES in Table \ref{tab:ablate}.

Individually, LoRA, L-Adapter, and S-Prefix outperform zero-shot CLIP by a large margin. 
The TPR task requires understanding both global and local relationships between textual descriptions and person image features. 
Specifically, S-Prefix (63.69\% R@1) improves attention by adding context-specific embeddings, surpassing the prompt-based submodule of CSKT (62.82\% R@1) by 0.86\%, validating its effectiveness.
LoRA (72.56\% R@1) adjusts attention weights to highlight subtle person features. 
L-Adapter optimizes the global feature distribution in LayerNorm, achieving superior performance on the coarse evaluation criterion R@10 with 93.81\%.
Hence, L-Adapter enhances global knowledge transferring, LoRA refines attention to capture local dependencies and features, and S-Prefix strengthens the prior relevant knowledge representation for TPR tasks.

\blue{To thoroughly validate the synergistic effectiveness of different submodules, we first integrate LoRA and L-Adapter (\emph{No.4}), which yields superior performance compared to their separate utilization (\emph{No.2} and \emph{No.3}), gaining an improvement by 0.92\% and 1.16\% on comprehensive retrieval criterion mAP.
Moreover, it is observed that the combination of S-Prefix and L-Adapter  (\emph{No.6}) works better on critical metrics R@1 and mAP where the combined result surpasses the effectiveness of individual component.
Integrating LoRA and S-Prefix also achieves superior performance on all metrics compared to individual component, demonstrating that these two submodules can work together to optimize MHA layer, enhancing the model's task-specific information while focusing on fine-grained features for attention.}

\blue{Finally,  the combination of the three submodules with L-Adapter, S-Prefix and LoRA (\emph{No.7}) surpasses other combinations on the most important retrieval metrics including R@1 (74.17 \%) and mAP (65.91 \%), This integration adjusts global features through L-Adapter, emphasizes fine-grained features in attention via S-Prefix and LoRA, and achieves effective synergistic functions within our three components.}

\blue{In overall, ablations on the three modules further validate the rationality of our architecture, as outlined in the theoretical analysis of Section \ref{ssec:Unified}.
The performance optimization of the three submodules shows no significant conflicts. 
Thus, their collaborative interaction results in better retrieval performance rather than single component across multiple metrics.}

\begin{table}[!tb]
\centering
\tabcolsep=3.5pt
\caption{Ablation study on each component of UP-Person.}
\vspace{-2mm}
\renewcommand\arraystretch{1.1} 
  \resizebox{0.48\textwidth}{!}{%
  \begin{tabular}{c|l|ccc|cccc}
  \toprule
  \multirow{2}{*}{No.} &\multirow{2}{*}{Methods} &\multicolumn{3}{c|}{Components} &\multicolumn{3}{c}{CUHK-PEDES}  \\ 
  \cline{3-9}
       &                                         &S-Prefix & LoRA     &L-Adapter      &R@1  &R@5  &R@10 & mAP \\ 
  \hline
  0    &Zero-shot CLIP                               &          &          &            &12.61	&27.08	&35.48 & 11.14   \\
  1    &+S-Prefix &     \checkmark     &          &       &  63.68 &  83.74 &  89.54 &57.21     \\
  2    &+LoRA                                       & &      \checkmark    &            & 72.56 & 89.26 & 93.18 & 64.96	     \\    
  3    &+L-Adapter                                     &          &          &\checkmark  &72.09 & 89.07 & \underline{93.81} & 64.73   \\
  4    &+LoRA   +L-Adapter                                   & &\checkmark & \checkmark  & \underline{73.54} & \underline{89.52} & 93.41& \underline{65.89} \\
5    &+S-Prefix +LoRA                                    &\checkmark &\checkmark &    & 72.95& 89.31& 93.70 & 64.96    \\
6    &+S-Prefix +L-Adapter                                    &\checkmark & & \checkmark  & 72.79 & 89.46 & 93.57 & 65.05  \\
  7    & \textbf{UP-Person (Ours)}                                  &\checkmark &\checkmark&\checkmark  & \textbf{74.17} & \textbf{89.70} & \textbf{93.88} & \textbf{65.91}  \\
  \bottomrule
  \end{tabular}%
  }
  \vspace{-2mm}
  \label{tab:ablate}
  \end{table}
  
\begin{figure*}[!tb]
  \centerline{\includegraphics[width=0.9\textwidth,height=0.62\textwidth]{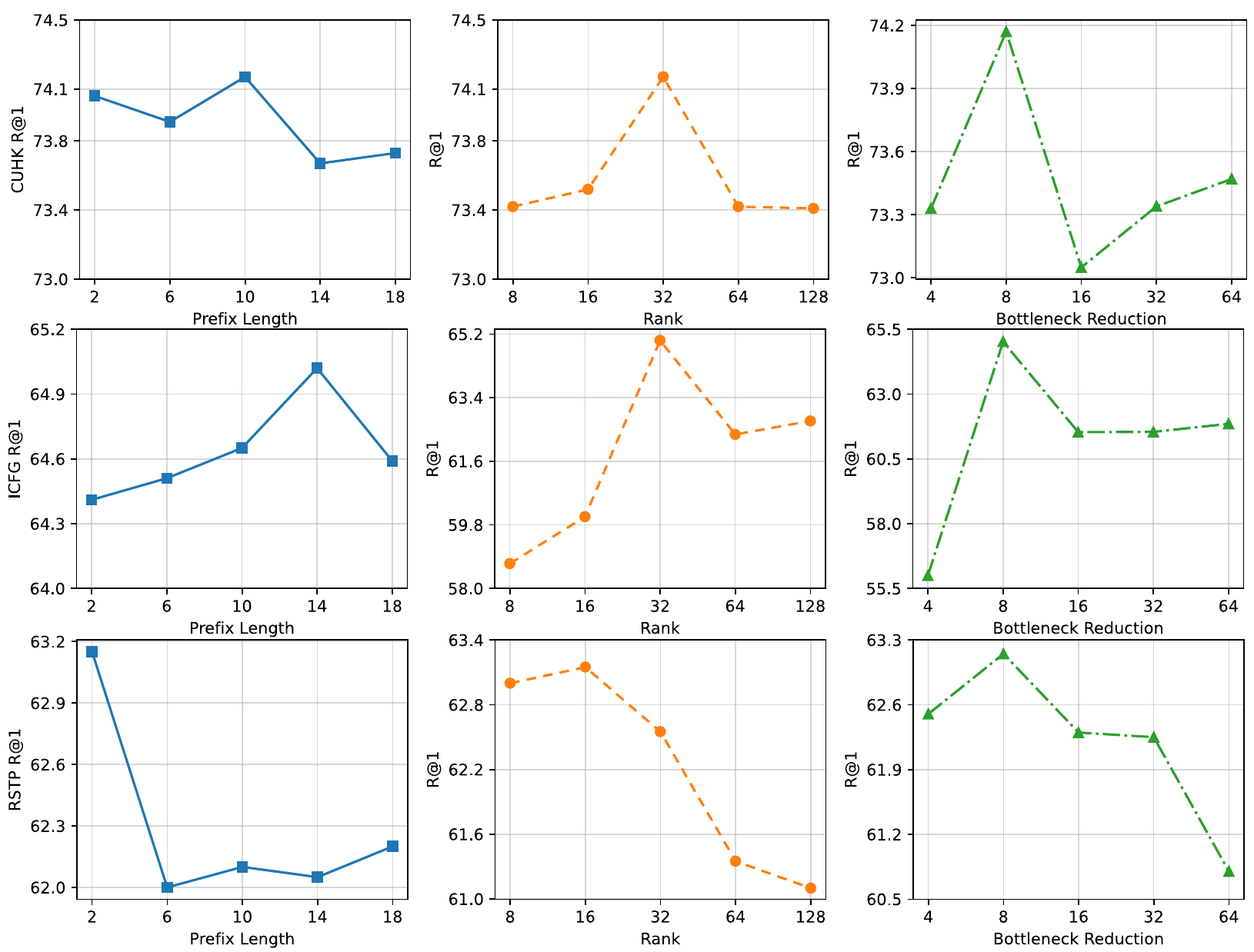}}
  \vspace{-2mm}
  \caption{\blue{\textbf{The analysis of hyper-parameters at R@1 on CUHK-PEDES, ICFG-PEDES and RSTPReid.}}}
  \label{fig:para}
\vspace{-3mm}
\end{figure*}

\textbf{Hyper-parameters.}
\blue{As shown in Figure \ref{fig:para}, to optimize performance across different datasets, we carefully select hyper-parameters based on dataset characteristics, where a grid search strategy is employed for each TPR dataset.
The bottleneck reduction $b$ of L-Adapter is consistently set to 8 across all three TPR datasets to balance performance and efficiency.
For LoRA, its rank $r$ is adjusted according to dataset size: CUHK-PEDES and ICFG-PEDES, being relatively larger datasets, use a rank of 32 to ensure sufficient representational capacity, whereas RSTPReid, with its smaller dataset size, benefits from a reduced rank of 16 to prevent overfitting while maintaining effective adaptation. 
Similarly, the length of S-Prefix $l$ is tailored to dataset-specific needs: CUHK-PEDES (10), ICFG-PEDES (14), and RSTPReid (2). 
This reflects that for small datasets, choosing a smaller $l$ helps control the model's parameter size and improves its performance on the limited data.}

\textbf{Ablations on LoRA and its variants.} \blue{ As shown in Tablse \ref{tab:lora_ablate}, we further analyze the ablation experiment on LoRA and its variants, \emph{i.e.,} DoRA \cite{liu2024dora}, and WoRA \cite{sun2024data} in our architecture.
We observe that the method incorporating the LoRA component outperforms those with DoRA and WoRA in overall metrics, both of which introduce additional parameters compared to LoRA.
Therefore, we ultimately choose LoRA submodule in our UP-Person.}

\begin{table}[!tb]
    \centering
    \caption{\blue{Ablation study on LoRA and its variants on CUHK-PEDES. The parameter configurations of other modules remain consistent across all three methods.}}
    \renewcommand\arraystretch{1.1} 
    \resizebox{0.45\textwidth}{!}{%
       \begin{tabular}{lccccc}
            \hline
            Method  & R@1  & R@5 & R@10 & mAP & Trainable \#Param (M) \\
            \hline
            UP-Person (LoRA \cite{hu2021LoRA})      &\textbf{74.17} & \textbf{89.70}& \textbf{93.88} &\textbf{65.91} & \textbf{7.42}\\
            UP-Person (DoRA \cite{liu2024dora})      & \underline{73.94} & \underline{89.31} & 93.47 & \underline{65.63} & \underline{7.45}\\
            UP-Person (WoRA \cite{sun2024data})       & 73.67 & 89.21 & \underline{93.62} & 65.54 & 7.45  \\
            \hline
        \end{tabular}
    }
      \vspace{-3mm}
    \label{tab:lora_ablate}
\end{table}

\subsection{Analysis of S-Prefix.}
Table \ref{tab:prefix} demonstrates the effectiveness of S-Prefix compared to vanilla prefix based on the CLIP backbone and UP-Person framework.
First, we take CLIP (S-Prefix) as an example to clarify the individual effect of $S_p$.
Compared to CLIP (Vanilla Prefix), CLIP (S-Prefix) gains a dramatic improvement of 5.64\% when $S_p=150$, demonstrating that integration $S_p$ can significantly improve retrieval performance.
With an increment of $S_p$, all metrics are enhanced, indicating that extending the attention values corresponding to the prefix part can effectively alleviate the problem of slow gradient changes.
However, excessively large $S_p$ such as $S_p=500$ leads to gradient exploding, and then the algorithm fails to converge.
In the unified framework which incorporates LoRA  and L-Adapter, UP-Person (S-Prefix) has a significant gain than UP-Person (Vanilla Prefix) by +0.53\% on R@1  when $S_p=10$.
We finally choose the appropriate parameter $S_p =10$ in our UP-Person.

\begin{table}[!tb]
    \centering
    \caption{Analysis of S-Prefix}
    \vspace{-2mm}
\renewcommand\arraystretch{1.1} 
  \resizebox{0.49\textwidth}{!}{%
    \begin{tabular}{llllll}
        \hline
        Prefix Type  & $S_{p}$ & R@1 & R@5 & R@10 & mAP\\
        \hline
        CLIP (Vanilla Prefix)   & - & 58.04 &  79.08 &  86.65 &  52.35  \\
        CLIP (S-Prefix)     & 10 & 61.91 & 82.46 & 88.61 & 55.75 \\
        CLIP (S-Prefix)     & 50 &  63.13 & 82.67 & 89.33 & 56.98 \\
        CLIP (S-Prefix)     & 100 &  63.63 & 83.84 & 89.86 & 57.37 \\
        CLIP (S-Prefix)     & 150 &  63.68 & 83.74 & 89.54 & 57.21  \\
        CLIP (S-Prefix)     & 300 & 63.08 & 82.81 & 89.23 & 56.73 \\
        CLIP (S-Prefix)     & 500 & - & - & - & - \\
        UP-Person (Vanilla Prefix) & -  &  73.64 &  89.13 &  93.50 & 65.70  \\
        UP-Person (S-Prefix)    & 10 & \textbf{74.17} & \textbf{89.70} & \textbf{93.88} & \textbf{65.91}  \\
        UP-Person (S-Prefix)    & 30 &  73.28 & 89.64 & 93.71 & \underline{65.74} \\
         UP-Person (S-Prefix)    & 50 & \underline{73.80} & \underline{89.69} & \underline{93.86} & 65.65   \\
        \hline
    \end{tabular}
    }
    $S_{p}$ is a salable factor to boost the attention related to the prefix part. \\
    \label{tab:prefix}
     \vspace{-2mm}
\end{table}

\subsection{Analysis of L-Adapter.}
Table \ref{tab:adapter} shows the effectiveness of LN-tuning \cite{qi2022parameter}, vanilla adapter and our L-Adapter in UP-Person, and further explores the sequential and parallel connection. 
As depicted in Figure \ref{fig:adapter}(a), Sequential Adapter is inserted behind MLP and MHA, which is the original adapter structure as in \cite{houlsby2019parameter}.
Sequential L-Adapter is connected behind layernorm in series.
Similarly, parallel adapters span specific network components which belong to vanilla CLIP, which are demonstrated in Figure \ref{fig:adapter}(c) and (d).
LN-tuning in Figure \ref{fig:adapter}(e) fine-tunes the gain (for scale operation) and bias (for shift operation) parameters  while keeping other parameters frozen.

We observe that the performance of parallel vanilla adapter is far lower than that of parallel L-Adapter (-1.17\% on R@1).
Meanwhile, it is obvious that both sequential and parallel L-Adapter significantly surpass two other vanilla adapters, which denotes that layernorm is extremely important in fine-tuning to TPR due to scale and shift parameters.
Compared to LN-tuning, parallel L-Adapter gains an overwhelming improvement by +1.64\% on R@1.
We speculate that unfreezing layernorm  will directly destroy the knowledge of the pre-trained CLIP. 

\textbf{Analysis on conflict mitigation.}
As shown in Table \ref{tab:conflict}, we compare our UP-Person (\emph{No.1}) and other three combinations of adapter including the mixed adapter with two types of adapters (\emph{No.2} and \emph{No.4}) and vanilla adapter (\emph{No.3}).

The experimental results reveal that UP-Person  with both MHA L-Adapter and MLP L-Adapter (\emph{No.1}) outperforms all other combinations in terms of all evaluation metrics: R@1, R@5, R@10, and mAP, showcasing superior retrieval performance.
The vanilla adapter configuration (\emph{No.3}) performs slightly worse than the L-Adapter configurations, with lower values for R@1, R@5, R@10, and mAP compared to \emph{No.1}. This indicates that the presence of an L-Adapter, particularly in both the MHA and MLP layers, is crucial for boosting performance.
Other mixed configurations (\emph{No.2} and \emph{No.4}) also underperform to our full L-Adapter configuration, further suggesting that L-Adapter is better suited for the unified structure. 
We speculate that the vanilla adapter, spanning multiple network layers (MHA or MLP), introduces conflicts with other PETL components in the optimization space, whereas the L-Adapter, which focuses solely on adjusting the global distribution through LayerNorm, avoids these spatial conflicts without entangling, and optimizes the structure more effectively.

\begin{table}[!tb]
    \centering
    \caption{Analysis of L-Adapter}
    \vspace{-2mm}
\renewcommand\arraystretch{1.1} 
  \resizebox{0.48\textwidth}{!}{%
    \begin{tabular}{lllll}
        \hline
        Adapter Type  & R@1 & R@5 & R@10 & mAP\\
        \hline
        UP-Person (LN-tuning)       &  72.53 & 88.87 & 93.29 & 64.72 \\
        UP-Person (Sequential Adapter)       & 72.90 & 88.65 & 93.26 & 65.15 \\
        UP-Person (Sequential L-Adapter)      &  \underline{73.59} & \underline{89.26} & \underline{93.32} & \underline{65.78} \\
        UP-Person (Parallel Adapter)       &  73.00  & 88.55  & 92.95  & 65.07 \\
        UP-Person (Parallel L-Adapter)         & \textbf{74.17} & \textbf{89.70} & \textbf{93.88} & \textbf{65.91}   \\
        
        \hline
    \end{tabular}
    }
    
    \label{tab:adapter}
\end{table}

\begin{table}[!tb]
    \centering
    \caption{Analysis of Conflict Phenomenon. }
\renewcommand\arraystretch{1.1} 
  \resizebox{0.48\textwidth}{!}{%
    \begin{tabular}{llllll}
        \hline
         No. & Adapter Type  & R@1 & R@5 & R@10 & mAP\\
        \hline
        1 & UP-Person (MHA L-Adapter, MLP L-Adapter)  &    \textbf{74.17} & \textbf{89.70} & \textbf{93.88} & \textbf{65.91}  \\
        2 & UP-Person (MHA L-Adapter, MLP Adapter)         &  71.77 & 88.89 & 92.92 & 64.68   \\
       3 & UP-Person (MHA Adapter, MLP Adapter)         &  73.00 & 88.55 & 92.95 & 65.07  \\
       4 & UP-Person (MHA Adapter, MLP L-Adapter)         &   \underline{73.67} & \underline{89.08} & \underline{93.49} & \underline{65.72}  \\
        
        \hline
    \end{tabular}
    }
    
    \label{tab:conflict}

\end{table}

\subsection{Analysis of loss functions.}

\begin{table}[!tb]
    \centering
    \caption{Analysis of loss functions.}
    \renewcommand\arraystretch{1.1}
    \resizebox{0.45\textwidth}{!}{%
        \begin{tabular}{lcccccc}
            \hline
            Loss    & R@1  & R@5  & R@10   & mAP & Trainable \#Param (M) $\downarrow$\\
            \hline
              SDM &  \textbf{74.17} & \underline{89.70} & \underline{93.87} & 65.91  & \textbf{7.4M}\\
              ITC &   72.04 & 88.37 & 93.36 & 64.71    & \textbf{7.4M} \\
              SDM + ITC  &   \underline{73.59} & 89.70 & 93.70 & 65.90 & \textbf{7.4M} \\

              SDM + ID   & \underline{73.59} & 89.62 & 93.65 & \underline{66.57} & \underline{13.1M} \\

             SDM + ID + MLM  &  73.44 & \textbf{89.86} & \textbf{93.93} & \textbf{66.64} &  52.3M\\

            \hline
        \end{tabular}} 
        \label{tab:loss_func}
\end{table}

\blue{Table \ref{tab:loss_func} provides ablation studies on various loss functions, including the ITC (Image-Text Contrastive) loss, ID loss (for person identity classification) and MLM loss (for masked implicit reasoning) based on IRRA \cite{jiang2023cross}. 
It shows that SDM achieves superior results in all metrics compared to ITC loss by +2.13\%, +1.33\%, +0.51\%, and +1.20\%, respectively. 
The overall Rank metrics with ID loss are lower than those of SDM alone, where ID loss  primarily focuses on classification and plays a supplementary role in TPR task.
The combination of SDM+ID+MLM improves R@5, R@10 and mAP metric due to its integration with the cross-attention module for implicit reasoning.
However, this slight gain from MLM comes at the cost of an dramatic increase in model size and trainable parameters. 
Therefore, UP-Person with SDM offers an optimal trade-off between complexity and performance without bells and whistles, and its simplicity provides a scalable foundational framework, making it easy to combine with different types of losses based on objectives. } 

            

\subsection{Domain Generalization Performance Comparison}
\begin{table*}[!htb]
    \centering
    \caption{Domain Generalization Performance Comparison} 
    \vspace{-2mm}
\renewcommand\arraystretch{1.1} 
  \resizebox{0.95 \textwidth}{!}{%
    \begin{tabular}{cc|llll|llll|llll}     
        \hline
        \multicolumn{2}{c|}{Domains}  &\multicolumn{4}{c|}{IRRA} &\multicolumn{4}{c|}{IRRA-CLIP} &\multicolumn{4}{c}{UP-Person}   \\ 
        \hline
        
         Source Domain & Target Domain  &  R@1 & R@5 & R@10 & mAP & R@1 & R@5 & R@10 & mAP & R@1 & R@5 & R@10 & mAP \\
        \hline
         CUHK   & ICFG & \underline{42.41} &  62.10 &  69.60 & 21.77  &37.46 & 57.59 & 66.13 & 19.44 &  \textbf{44.65} & 64.10 & 71.88 & 23.24  \\
         CUHK      & RSTP & \textbf{53.20} & 77.15 &  85.35 & 39.63  & 48.20 & 73.70 & 82.65 & 36.80 &\underline{52.40} & 77.30 & 84.45 & 39.70 \\
        ICFG     & CUHK & \underline{33.46} & 56.34 &66.33 & 31.57	&32.51 & 53.79 & 64.18 & 29.84 &\textbf{33.63} &56.99 & 67.09 & 31.21	\\
        ICFG      & RSTP &  \underline{45.30} & 69.30 & 78.80 & 36.82  & 40.25 &  64.30 &  74.45 & 31.41 & \textbf{45.80} & 71.00 & 79.40 & 36.43 \\
        \textbf{RSTP}      &  CUHK & \underline{32.75} & 55.26 & 65.81 & 30.29 &	31.53  &53.66 & 63.40 & 29.68 & \textbf{36.96} & 58.22 & 68.21 & 33.84 \\
         \textbf{RSTP}      &  ICFG & \underline{32.30} & 49.68 & 57.78 & 20.54&	29.03 & 46.87 & 55.54 & 18.26& \textbf{38.05} & 55.68 & 63.67 &22.89\\
        \hline
    \end{tabular}
    }
    \label{tab:generation}
    \vspace{-2mm}
\end{table*}

\begin{table}[!htb]
\centering
\tabcolsep=3.5pt
\caption{Comparison on Coarse-grained Datasets.}
\vspace{-2mm}
\renewcommand\arraystretch{1.1} 
  \resizebox{0.485\textwidth}{!}{%
  
  \begin{tabular}{l|c|c|ccc|ccc}
  \toprule
  \multirow{2}{*}{Methods} &\multirow{2}{*}{Ref} &  \multirow{2}{*}{\makecell[c]{Trainable\\\#Param}} &\multicolumn{3}{c|}{Flickr30k (1k test)} &\multicolumn{3}{c}{MSCOCO (5k test)}  \\ 
  \cline{4-9}
  & &  & R@1  &R@5  &R@10    &R@1  &R@5  &R@10 \\ 
  \hline

  IMEB  & TCSVT2024 & - & 64.00 & 88.0 & 92.80 & 44.90 & 74.60 & 84.00\\
   DSMD  & Arxiv2024 & 197M & 68.40 & 90.80 & 94.40 & 48.00 & 75.60 & 84.50\\
   BEIT-3  & CVPR2023 & 1.9B & 81.50 &95.60& 97.80 & \underline{67.20} & \textbf{87.70} & \textbf{92.80} \\
   BLIP-2  & ICML2023 & 1.2B & \textbf{89.70} & \textbf{98.10} &\textbf{98.90} & \textbf{68.30} & \textbf{87.70} & \underline{92.60} \\ 
  \textbf{UP-Person (Ours)}   & - &   7.4M&   \underline{83.92} & \underline{96.72} & \underline{98.58}   & 56.51& \underline{81.74} & 89.18   \\

  \bottomrule
  \end{tabular}%
  }
  \label{tab:coarse}
  \end{table}

In Table \ref{tab:generation}, we conduct domain generalization experiments between different datasets derived from CUHK-PEDES, ICFG-PEDES and RSTPReid.
We train models on the dataset of source domain and then transfer them to the target domain to verify the generalization performance of our method.
As we can see, UP-Person achieves comparable generalization performance with IRRA when source domain is CUHK-PEDES or ICFG-PEDES.
It further has an overwhelming advantage compared with IRRA when source domain is RSTPReid, which gains a significant improvement on R@1 by +4.21\% and +6.2\% when target domain is CUHK-PEDES and RSTPReid, respectively.
Moreover, it is obvious that UP-Person is superior than IRRA-CLIP (full-tuning) in all metrics by a large margin.

We conjecture that our method effectively alleviate overfitting compared to full-tuning-based methods since it merely fine-tunes a few parameters of simple PETL components.
This advantage is more pronounced when the training dataset is scarce.
We observe that the size of RSTPReid is much less than other datasets as elaborated in the beginning of Section \ref{sec:expe}.
Thus, full-tuning-based methods such as IRRA and IRRA-CLIP can cause the more severe overfitting when lacking of training data and further largely reduce generalization performance.
All results demonstrate the powerful generalization ability of our UP-Person.

\begin{figure}[!tb]
  \centerline{\includegraphics[width=0.5\textwidth,height=0.5\textwidth]{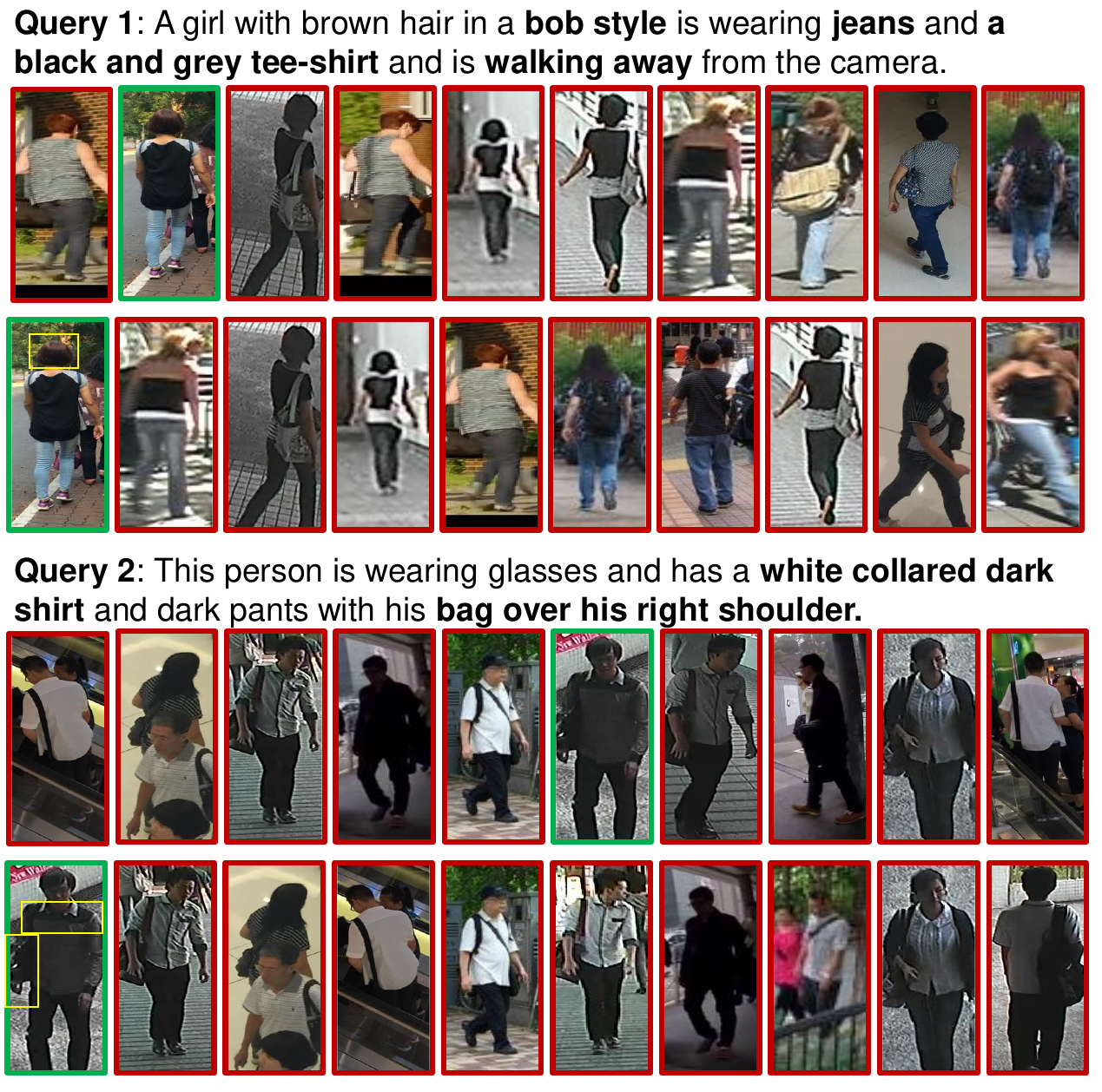}}
  \vspace{-2mm}
  \caption{\textbf{Comparison of top-10 retrieval results on CUHK-PDES by IRRA (the first row) and UP-Person (the second raw).} 
  The matched images are marked in green, and the unmatched ones are marked in red. The yellow box represents the subtle difference between correct and false retrieval results.}
  \label{fig:top10}
\vspace{-2mm}
\end{figure}
\subsection{Visualization}

Figure \ref{fig:top10} compares the top-10 retrieval results from the IRRA (the first row) and our proposed UP-Person (the second row). 
The matched and unmatched images are marked in green and red, respectively.
The yellow box denotes the key differentiated retrieval objects between correct and false retrieval.
It can be seen that UP-Person can retrieve the corresponding pedestrian images for a query text more accurately.
For example, IRRA cannot correctly recognize the phrase \emph{bob style} in Query 1, \emph{white collared dark shirt} and \emph{bag over his right shoulder} in Query 2, which represent fine-grained or unusual objects or words compared to common instances.
We infer that the full-tuning method IRRA, which only utilizes pre-trained parameters at initialization process, may lose part of the original abundant knowledge of vanilla CLIP during training.
If the model of IRRA does not fully ``see'' the relevant knowledge in training phase, inference when encountering the ``unseen'' objects may fail.
Overall, the visualization vividly demonstrates the effectiveness of UP-Person.

\subsection{Coarse-grained Text-to-Image Retrieval}
We utilize two datasets for the coarse-grained retrieval task: Flickr30K \cite{plummer2015flickr30k} and MSCOCO \cite{lin2014microsoft}. Unlike TPR datasets with single person object and fine-grained descriptions, these datasets contain various general objects and coarse-grained sentences. 
We follow the Karpathy split \cite{karpathy2015deep}, allocating 29K/1K/1K images for training, validation, and testing in Flickr30K, and 113K/5K/5K in MSCOCO.
Each image is annotated with five sentences.  

As shown in Table \ref{tab:coarse}, UP-Person on MSCOCO and Flicker exceeds other methods of similar scale  with millions of parameters by a large margin, such as IMEB \cite{li2024fast} and DSMD \cite{liang2024dynamic}.
Furthermore, BEIT-3 \cite{beit3} and BLIP-2 \cite{li2023blip} are larger multi-modal foundation models with more complex network and loss design for multiple vision-language tasks compared to CLIP.
Although this is an unfair comparison due to model scale, UP-Person with only 7.4M parameters outperforms BEIT-3 which has 1.9B trainable parameters on all metrics of Flicker30k.

\begin{table*}[!hbt]
    \centering
    \caption{Cross-Domain Generalization Performance Comparison} 
    \vspace{-2mm}
\renewcommand\arraystretch{1.1} 
  \resizebox{0.95 \textwidth}{!}{%
    \begin{tabular}{cc|llll|llll|llll}     
        \hline
        \multicolumn{2}{c|}{Domains}  &\multicolumn{4}{c|}{IRRA} &\multicolumn{4}{c|}{IRRA-CLIP} &\multicolumn{4}{c}{UP-Person}   \\ 
        \hline
        
         Source Domain & Target Domain  &  R@1 & R@5 & R@10 & mAP & R@1 & R@5 & R@10 & mAP & R@1 & R@5 & R@10 & mAP \\
        \hline
         Flickr30k   & CUHK & 19.40 & 37.39 & 46.44 & 17.22 & \underline{22.29} & 40.29 & 49.27 & 19.43 &  \textbf{23.36} & 42.35 & 52.57 & 20.45  \\
         Flickr30k      & ICFG & 8.18 & 19.20 & 26.35 & 2.90  &  \underline{9.63} & 21.68 & 29.06 & 3.51 & \textbf{9.90} & 22.36 & 29.75 & 3.77 \\
        Flickr30k     & RSTP & 20.50 & 44.45 & 57.15& 14.70	&\underline{21.45} & 45.20& 56.15 & 15.70 &\textbf{24.55} & 47.75 & 59.90 & 17.57	\\

        \hline
    \end{tabular}
    }
    \label{tab:coarse_generation}
    \vspace{-2mm}
\end{table*}
\blue{We conducted cross-domain generalization experiments, as shown in Table \ref{tab:coarse_generation}, where the source domain (coarse retrieval datasets) and the target domain (TPR dataset) exhibit significant differences.
We can observe that in the cross-domain generalization, our method also outperforms other approaches in all metrics, including all Rank metrics and mAP.}

In overall, UP-Person demonstrates considerable performance even on coarse-grained datasets, indicating that the unified framework has strong generalization capabilities.

\section{Deployment and Application}
\blue{The FLOPs of UP-Person is 13.78 GFLOPS, which indicates that its computational complexity is far below that of the current popular edge devices or mobile platforms, which typically range from 0.4 to 16 TOPs. This low computational complexity allows for efficient execution on these devices, making UP-Person suitable for real-time applications with limited computational resources.}

\blue{In terms of storage, the vanilla CLIP model can be shared across multiple pedestrian retrieval scenarios. Instead of storing a separate large foundation model for each specific scenario, UP-Person only occupies a few parameters tailored for each individual task.
This approach significantly reduces the storage space requirements on edge or mobile devices within a multi-scenario text-based person retrieval (TPR) system, optimizing resource usage without sacrificing performance.}


\section{Future Work}
\blue{UP-Person offers a simple yet effective foundational framework for cross-modal person retrieval, with significant potential for further development and broader applications.
One direction is advancing multimodal fusion by exploring finer-grained and more efficient interaction modules between different modalities besides foundational models. 
Additionally, incorporating diverse data sources, such as audio and visual modalities, could further broaden the range of information the model can utilize.
Audio data, such as sound cues, can provide context or details that text alone might not fully capture.
Visual modalities, like witness sketches, introduce a unique form of representation that can complement textual descriptions, particularly in scenarios where specific visual characteristics are challenging to describe precisely with words.
Another avenue is expanding the framework to support cross-lingual retrieval, enabling retrieval across multiple languages. 
Finally, investigating the scalability of UP-Person for larger datasets and real-time applications (\emph{e.g.,} through model compression and pruning) could offer valuable insights into its practical utility.}

\section{Conclusion}
In this paper, we propose a novel unified parameter-efficient transfer learning framework for text-based person retrieval based on CLIP and to fully transfer and explore knowledge within CLIP without bells and whistles.
This unified framework adjusts global features with our L-Adapter while capturing fine-grained features in attention with out optimized S-Prefix and LoRA.
The significant performance improvements across three widely-used person benchmark datasets and two general text-to-image retrieval datasets validate effectiveness, parameter-efficiency, and generalization of our method.
We hope our work can inspire future research on how to fully mine knowledge of the VLP models in a more effective and efficient PETL framework for TPR.

\bibliographystyle{IEEEtran}
\bibliography{bibtex}

\begin{IEEEbiography}[{\includegraphics[width=1in,height=1.25in,clip,keepaspectratio]{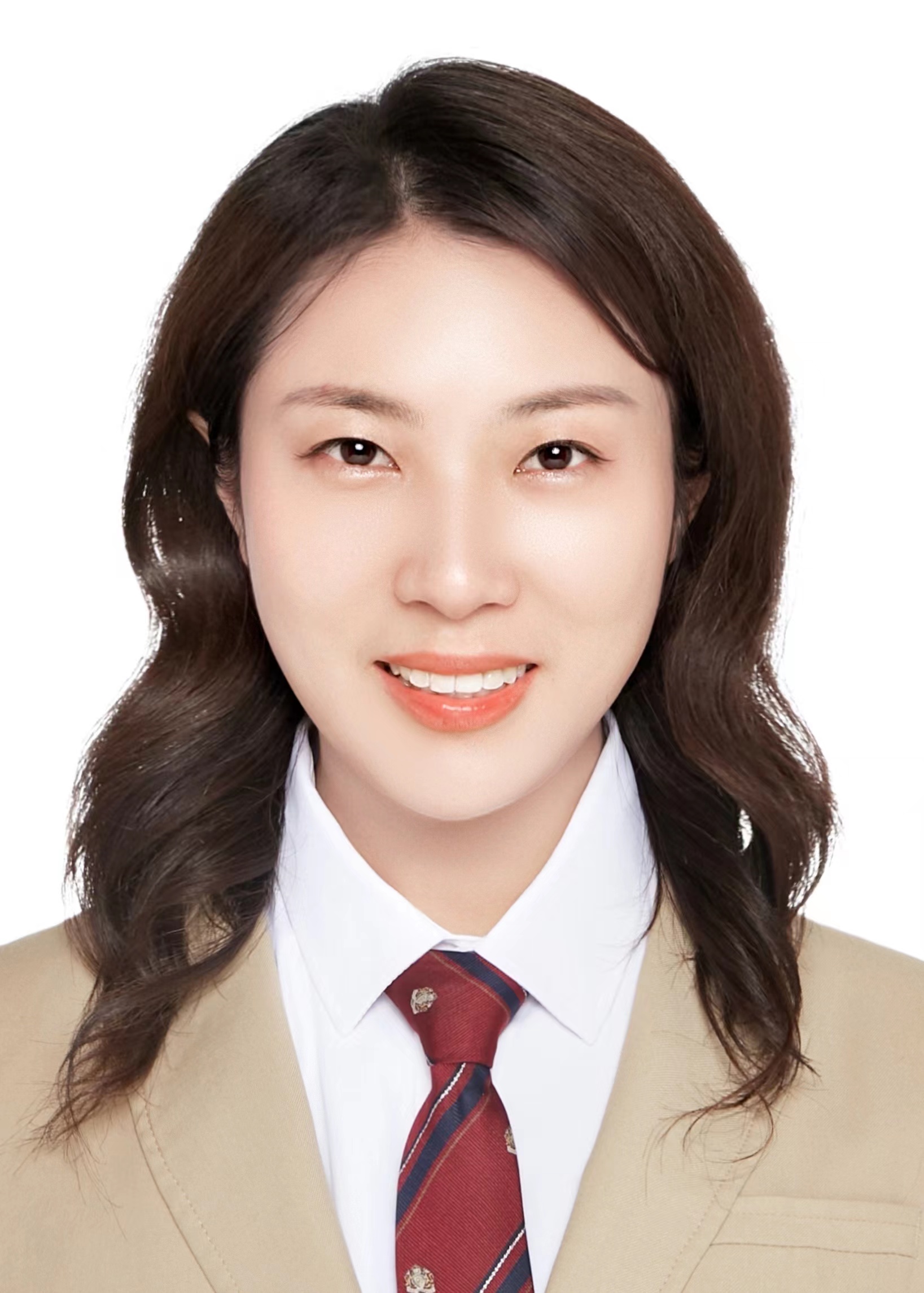}}]{Yating Liu} received the B.E.degree in communication engineering from Central South University, Changsha, China, in 2015, and the
M.S. degree in electronic and communication engineering from Tsinghua University, Beijing, China, in 2018. She is currently pursuing the Ph.D. degree with Tsinghua University and Peng Cheng Laboratory, Shenzhen, China. Her research interests include multimodal learning and cross-modal retrieval.
\end{IEEEbiography}

\begin{IEEEbiography}[{\includegraphics[width=1in,height=1.25in,clip,keepaspectratio]{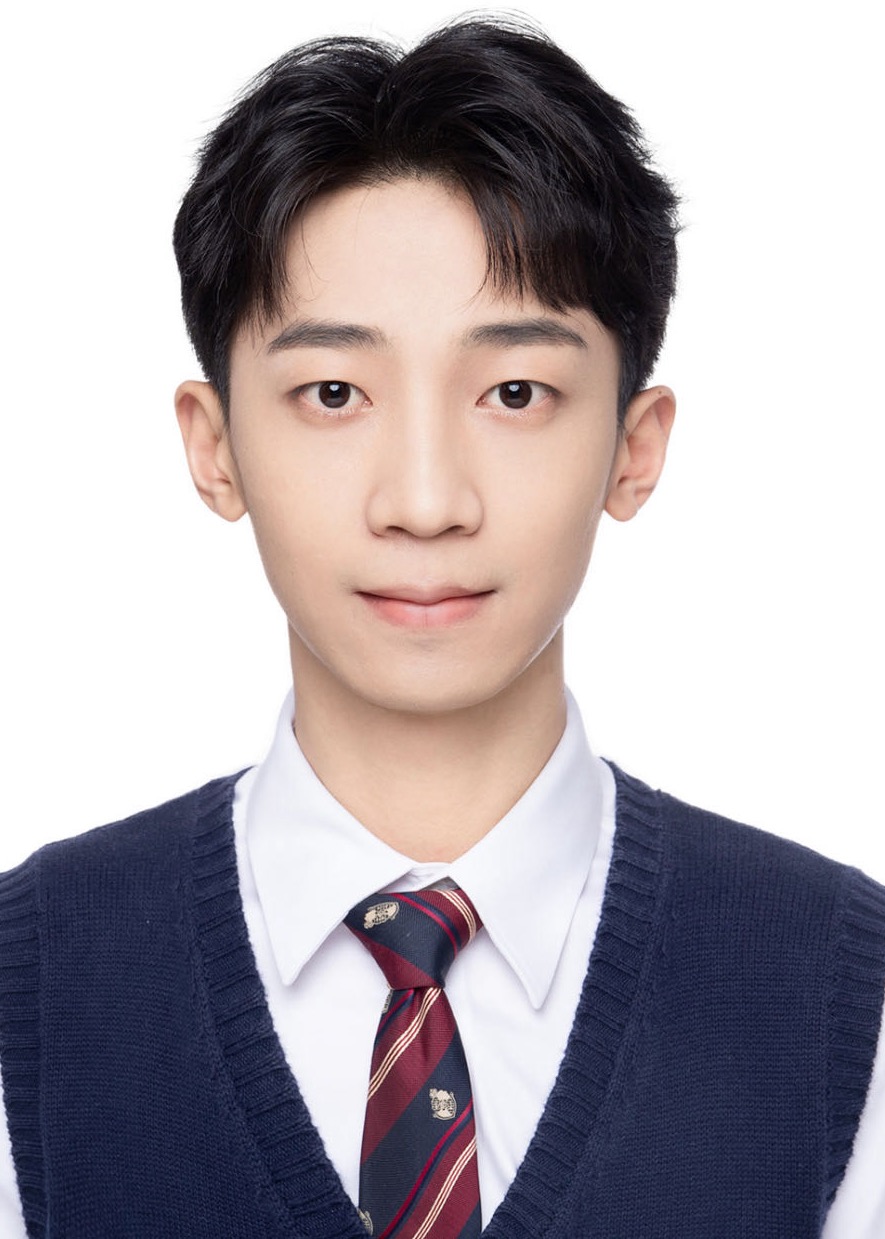}}]{Yaowei Li} received the B.E. degree in electronic and information Engineering from Yunnan University in 2022, and he is currently pursuing the Ph.D. degree from Peking University. His research interests include AIGC and generative diffusion model.
\end{IEEEbiography}

\begin{IEEEbiography}[{\includegraphics[width=1in,height=1.25in,clip,keepaspectratio]{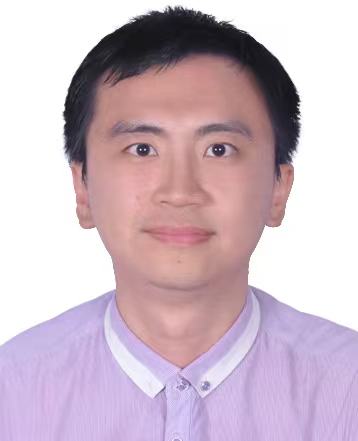}}]{Xiangyuan Lan} received the B.Eng. degree from the South China University of Technology, China, in 2012, and the Ph.D. degree from Hong Kong Baptist University, Hong Kong, in 2016. He was a Visiting Scholar with the University of Maryland, College Park, MD, USA, in 2015, and also a Visiting Researcher with the University of California at Merced, Merced, CA, USA, in 2017. He was a Post-Doctoral Research Fellow from 2016 to 2018 and then a Research Assistant Professor from 2018 to 2021, with Hong Kong Baptist University. He joined the Peng Cheng Laboratory, Shenzhen, China, in 2022, where he is currently an Associate Professor. He is a Ph.D. Supervisor affiliated with the Southern University of Science and Technology, China. He is currently an Associate Editor of Signal, Image and Video Processing and Frontiers in Signal Processing. His current research interests include pre-trained foundation models and their applications in multi-modal perception.
\end{IEEEbiography}

\begin{IEEEbiography}[{\includegraphics[width=1in,height=1.25in,clip,keepaspectratio]{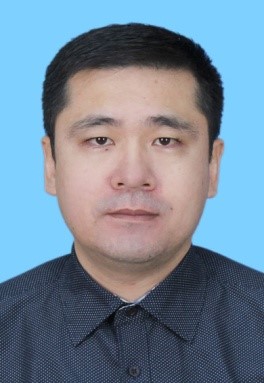}}]{Wenming Yang(IEEE Senior Member)} received his Ph.D. degree in information and communication engineering from Zhejiang University in 2006. He is an associate professor in Shenzhen International Graduate School/Department of Electronic Engineering, Tsinghua University. His research interests include image processing, computer vision, pattern recognition, deep learning and their applications.
\end{IEEEbiography}

\begin{IEEEbiography}[{\includegraphics[width=1in,height=1.25in,clip,keepaspectratio]{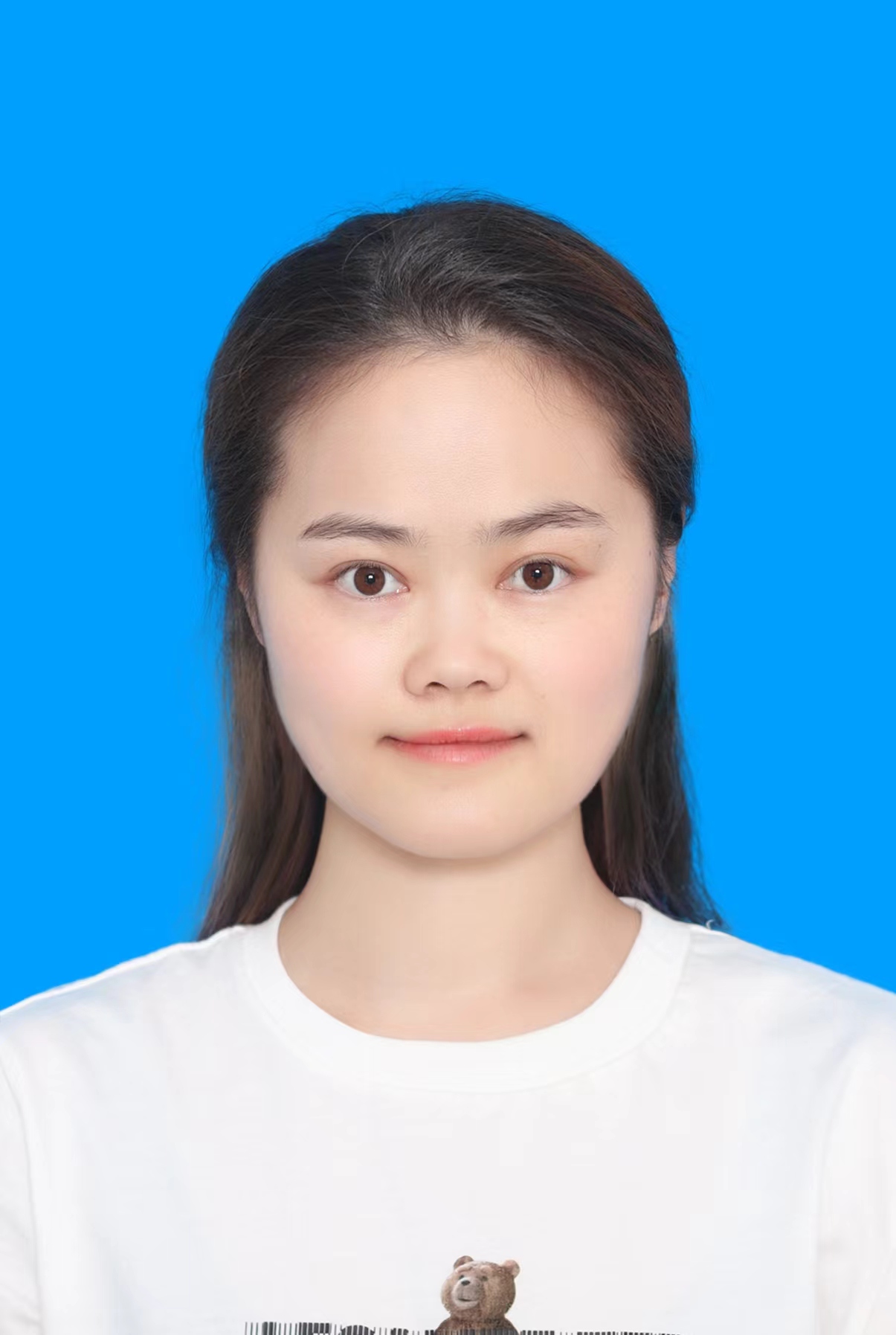}}]{Zimo Liu} received the Ph.D degrees in information and communication engineering in Dalian University of Technology, Dalian, China. She is currently a Post Doc in Peng Cheng Laboratory. Her current research interests include computer vision, model/feature compression and multi-modal learning.
\end{IEEEbiography}

\begin{IEEEbiography}[{\includegraphics[width=1in,height=1.25in,clip,keepaspectratio]{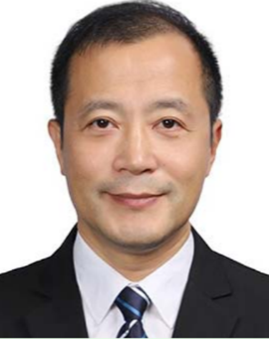}}]{Qingmin Liao  (IEEE Senior Member)} received the B.S. degree in radio technology from the University of Electronic Science and Technology of China, China, in 1984, and the M.S. and Ph.D. degrees in signal processing and telecommunications from the University of Rennes 1, France, in 1990 and 1994, respectively. He is currently a Professor with the Shenzhen International Graduate School/Department of Electronic Engineering, Tsinghua University. His research interests include image/video processing, analysis, biometrics, and their applications.
\end{IEEEbiography}

\vfill

\end{document}